\documentclass{article}

\usepackage{microtype}
\usepackage{graphicx}
\usepackage{booktabs} 
\usepackage{amsmath,amsfonts,bm}

\usepackage{hyperref}

\usepackage[accepted]{icml2021}

\usepackage{cleveref}
\crefformat{section}{\S#2#1#3} 
\crefformat{subsection}{\S#2#1#3}
\crefformat{subsubsection}{\S#2#1#3}

\newcommand{\ie}{{\em i.e.,}\ }
\newcommand{\eg}{{\em e.g.,}\ }
\newcommand{\Eg}{{\em E.g.,}\ }
\newcommand{\wrt}{\emph{w.r.t.}\ }

\definecolor{mypink3}{cmyk}{0, 0.7808, 0.4429, 0.1412}

\newcommand{\Ni}{({\em i})~}
\newcommand{\Nii}{({\em ii})~}

\def\eqref#1{equation~\ref{#1}}

\def\1{\bm{1}}

\def\vo{{\bm{o}}}
\def\vp{{\bm{p}}}

\def\vx{{\bm{x}}}
\def\vy{{\bm{y}}}

\DeclareMathAlphabet{\mathsfit}{\encodingdefault}{\sfdefault}{m}{sl}
\SetMathAlphabet{\mathsfit}{bold}{\encodingdefault}{\sfdefault}{bx}{n}

\def\gL{{\mathcal{L}}}

\def\gP{{\mathcal{P}}}

\def\gV{{\mathcal{V}}}

\def\sC{{\mathbb{C}}}

\def\sS{{\mathbb{S}}}

\def\sV{{\mathbb{V}}}

\newcommand{\R}{\mathbb{R}}

\newcommand{\softmax}{\mathrm{softmax}}

\usepackage{caption}
\usepackage{wrapfig}
\usepackage{subcaption}
\usepackage{comment}
\usepackage{hyperref}
\usepackage{url}
\usepackage{bbm}
\usepackage{mathtools}
\usepackage{booktabs}  
\usepackage{siunitx}

\icmltitlerunning{Straight to the Gradient: Learning to Use Novel Tokens for Neural Text Generation}

\begin{document}

\twocolumn[
\icmltitle{Straight to the Gradient: \\ Learning to Use Novel Tokens for Neural Text Generation}



\icmlsetsymbol{equal}{*}

\begin{icmlauthorlist}
\icmlauthor{Xiang Lin}{ntu}
\icmlauthor{Simeng Han}{ntu}
\icmlauthor{Shafiq Joty}{ntu,sf}

\end{icmlauthorlist}

\icmlaffiliation{ntu}{Nanyang Technological University, Singapore}
\icmlaffiliation{sf}{Salesforce Research Asia, Singapore}

\icmlcorrespondingauthor{Xiang Lin}{linx0057@e.ntu.edu.sg}

\icmlkeywords{Machine Learning, ICML}

\vskip 0.3in
]



\printAffiliationsAndNotice{}  


\begin{abstract}
Advanced large-scale neural language models have led to significant success in many language generation tasks. However, the most commonly used training objective, Maximum Likelihood Estimation (MLE), has been shown problematic, where the trained model prefers using dull and repetitive phrases. In this work, we introduce ScaleGrad, a  modification straight to the gradient of the loss function, to  remedy the degeneration issue of the standard MLE objective. By directly maneuvering the gradient information, ScaleGrad makes the model learn to use novel tokens. Empirical results show the effectiveness of our method not only in open-ended generation, but also in directed generation tasks. With the simplicity in architecture, our method can serve as a general training objective that is applicable to most of the neural text generation tasks.
\end{abstract}

\section{Introduction}

Text generation has been one of the most important research problems in natural language processing (NLP). Thanks to the advances in neural architectures, models are now capable of generating texts that are of better quality than before \citep{brown2020language}. However, despite the countless efforts that have been made to improve neural architectures, models trained with the standard {Maximum Likelihood Estimation} (MLE) objective are known to prefer generating dull and highly repetitive texts. For instance, in \emph{open-ended generation} tasks, such as story continuation or open dialogue generation, it has been observed that even with large pre-trained models like GPT-2 \citep{radford2019language}, high frequency tokens largely dominate the generation \citep{welleck2019neural,holtzman2019curious}. Similar observation has been reported in \emph{directed generation} tasks such as summarization \citep{see-etal-2017-get}, image captioning \citep{melas-kyriazi-etal-2018-training,Wang_Image2019} and machine translation \citep{tu-etal-2016-modeling,stahlberg-byrne-2019-nmt}.

The methods proposed to solve the degeneration issues with neural text generation can be primarily categorized into two groups: \Ni \emph{training} based methods, which include incorporating auxiliary losses \citep{see-etal-2017-get, welleck2019neural, li-etal-2020-dont} and coverage vector \citep{see-etal-2017-get,tu-etal-2016-modeling};
\Nii \emph{decoding} based methods, such as stochastic beam search \citep{kool2019stochastic}, top-$k$ sampling \citep{fan2018hierarchical}, nucleus or top-$p$ sampling \citep{holtzman2019curious}, and {inverse probability weighting \citep{zhang2021improving}}.

Though decoding based methods, in particular nucleus and top-$k$  sampling, perform well in practice in open-ended generation tasks, significantly reducing the degeneration problem, they do not address the fundamental modeling issue that the token-level probabilities produced by the neural model are problematic \citep{welleck2019neural}. {In addition, our experiments demonstrate that sampling methods also fail to generate high-quality texts in directed generation tasks such as abstractive text summarization.}

In this work, based on the known observation that the text generation models trained with MLE objective tend to generate repetitive tokens or phrases, we introduce a novel method called {ScaleGrad} for neural text generation training, by directly maneuvering the gradients to make the model learn to use novel tokens during training. Our method lies in {the} training based group, which aims to address the \textbf{fundamental modeling} problem, that is, the token-level distribution predicted by the generation model.

In a concurrent work, \citet{wang2020contextual} introduce a temperature scaling approach called Contextual Temperature to improve general language modeling. In this approach,  the temperature value in the $\softmax$ function is  parameterized by a neural network that is jointly trained with the main model. Though the objective of their work is not explicitly related to text degeneration, their analysis shows temperature scaling essentially changes the gradient updates that each token receives during training, which further motivates our work.



We conduct experiments with different neural architectures including LSTM \citep{hoch1997lstm} and Transformer \citep{vaswani2017selfatten} across different tasks in opened-ended and directed text generation. Through extensive analysis we demonstrate that {ScaleGrad} consistently improves the generation quality according to both human evaluation and automatic metrics. Compared to other training based methods, {ScaleGrad} is architecturally simpler and easier to fit into current neural models {(\cref{subsec:method})}, {while possessing a wide applicability to different text generation tasks
(\cref{subsec:directgen} and \cref{subsec:analysis on directed}).} The source code is available at \url{https://github.com/shawnlimn/ScaleGrad}.

\section{Background}

\subsection{Neural text generation}
The NLP tasks involving text generation can be broadly categorized into two types: \textit{directed generation} and \textit{open-ended generation} \citep{holtzman2019curious}. In the former case, the output text can be seen as a constrained transformation of the input. Examples include text summarization,  machine translation, and image captioning. In the latter case, the input context only provides a certain degree of constraints such that the model is allowed to generate the following texts with a considerable degree of freedom. Story/text continuation and dialogue generation fall in this category.

Neural models frame text generation tasks as some form of conditional language modeling, which is typically trained to maximize the log likelihood (equivalently, minimize the negative log likelihood) of the training data. The \textit{Maximum Likelihood Estimation} or MLE objective for an input-output pair $(\vx, \vy)$ can be expressed as follows.  

\begin{equation}
\small 
    \mathcal{L}_{\text{MLE}} = -\sum_{t=1}^{T} \log \gP_\theta(y_t|y_{<t}, \vx) \label{eq:mle}
\normalsize
\end{equation}
\noindent where $\theta$ denotes model parameters, $T$ is the length of the output sequence $\vy$, and $\vx$ is the task-specific input condition, \eg source document in summarization, image in image captioning, conversation history in dialogue generation and $\emptyset$ in text continuation. \textit{Teacher Forcing} \citep{williams1989teacherforcing}, {where current step's target token is passed as the next input to the decoder rather than the predicted token, is usually used to train the models for faster convergence.}

\paragraph{Degeneration} 
Degeneration has been a key problem in neural text generation models for open-ended tasks, {where the model generates texts that are repetitive, overly generic (dull), incoherent and gibberish. It can happen at different levels of granularity -- token, phrase, sentence and paragraph.} The problem has not been mitigated even with large-scale pre-trained models like GPT-2 Large \citep{radford2019language,holtzman2019curious}. 
Degeneration has also been observed in directed generation tasks even though the output in these tasks is confined by the input. For instance, in text summarization, most of the advanced models such as BertSum \citep{liu-lapata-2019-text}, BART \citep{lewis2019bart} and ProphetNet \citep{yan2020prophetnet} make use of tri-gram blocking \citep{Paulus2018ICLR} within beam search to remove duplicate trigrams during decoding, which improves the generation quality  in terms of the automatic metric. This implies that even with involvement of large-scale pre-trained models,  degeneration still exists. Similar issues have been reported in machine translation \citep{koehn-knowles-2017-six,stahlberg-byrne-2019-nmt}, image-description generation \citep{melas-kyriazi-etal-2018-training,Wang_Image2019} and next utterance generation in conversations \citep{jiang2020tldr}.

\subsection{Combating neural text degeneration}

Out of the methods proposed to tackle neural text degeneration, 
top-$k$ sampling \citep{fan2018hierarchical} and nucleus sampling \citep{holtzman2019curious}  stand out as representatives of decoding based methods and unlikelihood training \citep{welleck2019neural} as a representative training based method.
During each decoding step, nucleus and top-$k$ sampling use {different functions} 
to filter the candidate tokens, thus reformalizing the probability distribution. Then they sample the next token from the new distribution instead of maximizing the actual likelihood. Randomness brought by these sampling methods reduces duplicate tokens in the output. However, decoding strategy solely does not solve the underlying modeling problem with MLE training, as pointed out by \citet{welleck2019neural}. Our analysis in \Cref{subsec:analysis on directed} also reveals that sampling methods fail to generate high-quality texts in directed generation tasks.

To address the issue with MLE, \citet{welleck2019neural} propose the neural unlikelihood (UL) training method. During training, at each decoding step $t$, UL adds an auxiliary loss to the original cross entropy loss as follows.
\begin{equation}
\small 
    \mathcal{L}^t_{\text{UL}} = - \underbrace{\log  \gP_\theta(y_t|y_{<t})}_\text{MLE} - \alpha \sum_{c \in \sC^t} \underbrace{ \log (1 - \gP_\theta(c|y_{<t}))}_\text{UL} \label{eq:ul-loss} 
\normalsize
\end{equation}
\noindent where $\alpha$ is a hyper-parameter and $\sC^t$ is the set of \emph{negative tokens} at decoding step $t$, which is constructed by previous context tokens that are not the current token, $\sC^t = \{y^1, \ldots, y^{t-1}\} \setminus y^t$. The auxiliary UL loss decreases the total loss based on the ``unlikely'' probabilities of negative tokens, thus implicitly reducing the probability assigned to the repetitive tokens. UL training targets at improving the underlying modeling problem, which accords with our goal. Therefore, we mainly compare our method with UL training.\footnote{\citet{welleck2019neural} also propose a sequence-level UL. Since our work focuses on token-level modeling, we compare with their token-level UL training in this work.} We discuss how our method is different from UL training from the gradient perspective in \cref{subsec:issueoful}.

\section{Methodology: learning to use novel tokens}

Training {a text generation model} with MLE objective treats each token in the gold (ground truth) sequence equally. It has been shown that with this approach, the model 
exhibits the tendency to
generate repetitive tokens/phrases during inference \cite{welleck2019neural,holtzman2019curious}. To mitigate this degeneration problem, we argue that the model should focus more on  \textit{learning to use novel tokens}, rather than treating all the tokens in a sequence equally. {Our main idea is to maintain a dynamic list of novel tokens at each decoding step during training and encourage the model to learn to use tokens from this list for generation.}

Formally, let $\vy = (y^1, \ldots, y^t, \ldots, y^T)$ be the ground-truth (target) token sequence that the model is learning to generate in an auto-regressive manner, one token at a time. At time step $t$, we define the token $\Tilde{y}^t_{i}$ in the vocabulary $\sV$ as a \textbf{novel token}, if $\Tilde{y}^t_i$ has not been generated  before, \ie\ $\Tilde{y}^t_i \notin$ $\{y^1, \ldots, y^{t-1}\}$. By the definition, we have a dynamic set of novel tokens $\sS_{\text{novel}}^t \subseteq \sV $ at each decoding step $t$ in training, which shrinks over time as new tokens are observed in the ground-truth sequence (see Appendix~\ref{append:novelsetfigure} for an illustration). Note that the set of \emph{non-novel} tokens at each step (\ie\ $\sV \setminus \sS_{\text{novel}}^t$) is equivalent to the set of negative tokens $\sC^t$ in UL (Eq. \ref{eq:ul-loss}) except that it may contain the current target token $y^t$, if it was observed before. To encourage the model to focus on learning to use novel tokens, we propose an architecturally-simple yet effective method that can fit into most of the {auto-regressive} generation models. Our method, requiring no carefully-designed components, is derived directly from the gradient analysis of the loss function.

\subsection{Gradient analysis for MLE training}
\label{subsec:mle_gradient}
Let us first consider the gradient analysis of the model trained with MLE. Let $\vo^t \in \R^{|\sV|}$ denote the pre-softmax scores (\ie\ logits) over the vocabulary at time step $t$, where $o_i^t$ is the score for the token with index $i$. %
Similarly, let 
$p_k^t = [\softmax(\vo^t)]_k$
represent the probability of the ground truth token with index $k$ in the vocabulary. The partial derivative of the MLE objective (Eq. \ref{eq:mle}) at time step $t$  with respect to the logit $o_i^t$ can be shown as (omitting $t$ and `$\text{MLE}$' subscript for simplicity):
\begin{equation}
    \nabla_{o_i} \mathcal{L}
    = \frac{\partial \mathcal{L}}{\partial p_k} \cdot \frac{\partial p_k}{\partial o_i}
    = p_i - \mathbbm{1}(i = k) 
    \label{eq:partial}
\normalsize
\end{equation}
where $p_i= [\softmax(\vo)]_i$ and {$\mathbbm{1}(\cdot)$ is the Indicator function} (derivation is given in Appendix \ref{append:derivations}). Specifically, the gradient of the loss \wrt the ground truth token logit $o_k$ is $(p_k - 1)$ and for any other token logit $o_i$ is $p_i$.  
As the gradient-based optimization proceeds, the gradient converges to $\epsilon$, a number that is close enough to $0$. Another interpretation is that the gradient of the loss is supposed to be close to $0$ around a (local) minimum. Therefore, to reach the minimum, or to make the gradient close to $0$, the model would try to  increase the probability of ground truth token $p_k$ and reduce the probability of non-ground truth token $p_i$ in the MLE training.

From Eq. \ref{eq:partial}, it is clear that the gradient that every token $o_i \in \sV$  receives is directly related to its generation probability $p_i$. Therefore, we hypothesize that directly manipulating the generation probabilities of tokens, thereby controlling their gradients, can help us achieve our goal, which is to train the model so that it is encouraged to use novel tokens.

\subsection{Our method: {ScaleGrad}}
\label{subsec:method}

To encourage the model to learn to use novel tokens for generation, we can control the gradient to force the model to either increase the probability of novel tokens or decrease the probability for non-novel tokens. Based on this basic idea, we propose an effective training method keeping it in the simplest form. Specifically, at each decoding step of training, we re-normalize the $\softmax$ output (the probability distribution over the vocabulary) in a way such that the model is {informed of the current set of novel tokens and encouraged to use them.} Assuming that ${\vp}^t$ is the $\softmax$ output at step $t$ and $\sS^t_{\text{novel}}$ is the corresponding set of novel tokens at that step,
we re-compute the probability distribution as follows (again omitting $t$ for notational simplicity): 
\begin{equation}
        \Tilde{p}_i = 
        \begin{dcases}
        \frac{\gamma \cdot {p}_i}{ \sum_{j=1}^{|\sS_{\text{novel}}|} \gamma \cdot p_j + \sum_{j=1}^{|\sV'|} p_j },  \quad \text{if } i \in \sS_{\text{novel}} \\ 
        \frac{{p}_i}{ \sum_{j=1}^{|\sS_{\text{novel}}|} \gamma \cdot p_j + \sum_{j=1}^{|\sV'|} p_j },  \quad \text{otherwise}
        \end{dcases}
        \label{eq:renorm}
\normalsize
\end{equation}
where $\sV' = \sV \setminus \sS^t_{\text{novel}}$ is the non-novel tokens set at step $t$ and $\gamma \in (0,1)$ is the only hyper-parameter in our method that controls to what degree we want to encourage the model to focus on novel tokens; a smaller value of $\gamma$ incurs more aggressive push for using novel tokens.

The effect of the above change is that we directly re-scale the generation probability (after re-normalization) of the tokens. For $i \in \sS_{\text{novel}}$, {the effective probability becomes} $\Tilde{p}_i = \lambda_i \cdot {p}_i$ with $\lambda_i \in (0,1)$, and for $i \notin \sS_{\text{novel}}$, {the effective probability becomes} $\Tilde{p}_i = \alpha_i \cdot {p}_i$  with $\alpha_i > 1$.\footnote{Note that $\lambda_i$ and $\alpha_i$ are functions of $p_i$ rather than constant numbers. \Eg $\lambda_i =  \gamma / (\sum_{j=1}^{|\sS_{\text{novel}}|} \gamma \cdot p_j + \sum_{j=1}^{|\sV'|} p_j$). } Since $\lambda_i \cdot {p}_i$ and $\alpha_i \cdot {p}_i$ are new re-normalized probabilities, they both are  naturally bounded in $[0,1]$. 
{Consequently, assuming that the ground truth  token is indexed at $k$, the modified loss function at step $t$ for our proposed method becomes:}
\begin{equation}
\begin{split}
{\gL_{\text{SG}} = - \sum_{i=1}^{|\sV|} \mathbbm{1}(i = k)}  \Big[ &\mathbbm{1}(i \in \sS_{\text{novel}}) \log (\lambda_i \cdot {p}_i)  \\ 
&+ \mathbbm{1}(i \notin \sS_{\text{novel}}) \log (\alpha_i \cdot {p}_i) \Big]
\end{split}
\label{eq:sg-loss}
\normalsize
\end{equation}
The gradient for each token has been changed to\footnote{Derivation is given in Appendix \ref{append:derivations}.}:
\begin{equation}
\begin{aligned}
        \nabla_{o_i}\mathcal{L} &= \Tilde{p}_i - \mathbbm{1}(i = k) \\
        &= 
    \begin{dcases}
    \lambda_i \cdot {p}_i - 1, \quad \text{if } i = k \text{ and } i \in \sS_{\text{novel}}\\
    \alpha_i \cdot {p}_i - 1, \quad  \text{if } i = k \text{ and } i \notin \sS_{\text{novel}}\\
    \lambda_i \cdot {p}_i, \quad \quad \ \ \ \text{if } i \neq k \text{ and } i \in \sS_{\text{novel}}\\
    \alpha_i \cdot {p}_i, \quad \quad \ \ \ \text{if } i \neq k \text{ and } i \notin \sS_{\text{novel}}\\
    \end{dcases} 
    \normalsize
\end{aligned}
\label{eq:sgcases}
\end{equation}
We now discuss why this re-scaling encourages the model to use novel tokens. As mentioned, during training {with each gradient update} the model tries to decrease the gradient norm to $0$ to reach a local minimum. First, for a ground truth token (\ie $i = k$), if it is also a novel token, the gradient norm $|\lambda_i \cdot {p}_i - 1|$ is pushed away from $0$ so that the model has to learn to increase the probability ${p}_i$ further to reduce the gradient norm; if it is not a novel token, $|\alpha_i \cdot {p}_i - 1|$ is pushed slightly closer to $0$, which still makes the model learn to predict the ground truth but with a relatively lower strength. For non-ground truth tokens (\ie $i \neq k$), when it is not a novel token, $|\alpha_i \cdot {p}_i|$ increases the gradient norm so that the model learns to assign much lower probability ${p}_i$ to reduce it. Similarly, when the token is novel but not a ground truth token, the resulting gradient norm $|\lambda_i \cdot {p}_i|$ becomes smaller, for which the model only moderately learns to decrease the probability ${p}_i$ to reduce the norm further.

\begin{figure}
\vspace{-0.8em}
    \centering
    \includegraphics[width=0.50\textwidth]{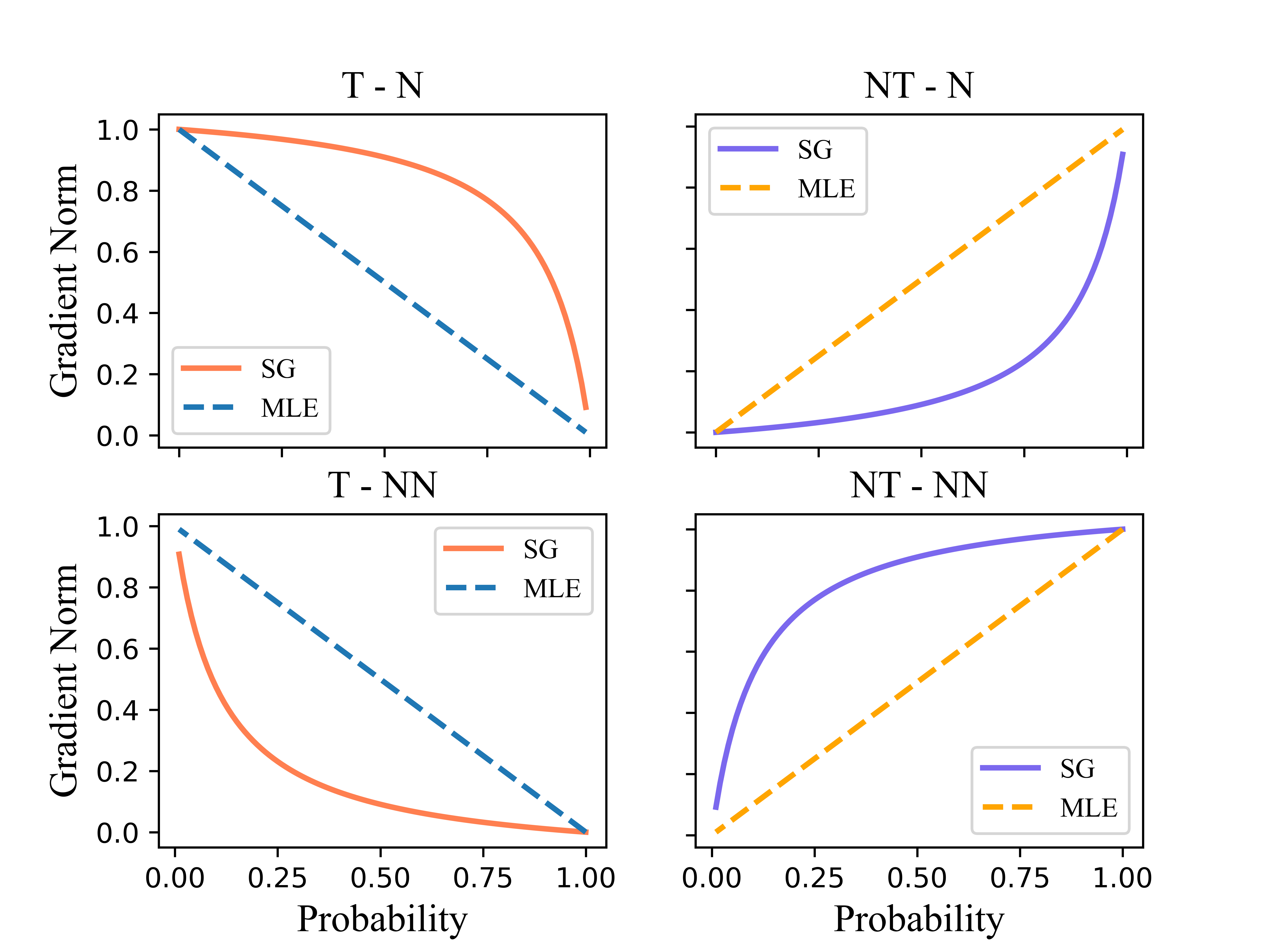}
    \caption{Illustration of the gradient norm for ScaleGrad and MLE. \textsc{T-N} denotes the \textbf{T}arget (ground truth) - \textbf{N}ovel token, \textsc{T-NN} denotes the \textbf{T}arget - \textbf{Non}-\textbf{N}ovel token, \textsc{NT-N} denotes the \textbf{Non}-\textbf{T}arget - \textbf{N}ovel token and \textsc{NT-NN} denotes the \textbf{Non}-\textbf{T}arget - \textbf{Non}-\textbf{N}ovel token. }
    \label{fig:gradient}
\end{figure}

{To give more insights, Figure~\ref{fig:gradient} plots the gradient norm for a toy example with two tokens, one of which is a novel token (\ie\ $|\sV| = 2$, $|\sS_{\text{novel}}| = 1$).} The dash lines represent the gradient information for MLE training, \ie $|p_i - \mathbbm{1}(i=k)|$. We can see how ScaleGrad scales the gradient dynamically for different types of tokens. For instance, for a target token belonging to the novel token set (\textsc{T-N}), its gradient norm has been scaled to a larger value compared to MLE, rendering that the model needs to learn to assign even higher probability to the target token to decrease the gradient. Similarly, for a non-target token that is not a novel token  (\textsc{NT-NN}), the scaled gradient makes the model assign even lower probability to the token in order to decrease the gradient. Moreover, the monotonic relation between the probability and the gradient norm guarantees that the model still learns to predict target tokens and reject non-target tokens, but in more dynamic degrees of strength.

\subsection{Comparison with unlikelihood training}
\label{subsec:issueoful}

We now analyze UL from the perspective of its gradients and compare with ours. The gradient of the UL loss (Eq. \ref{eq:ul-loss}) with a single negative token {(\ie\ $|\sC^t| = 1$)} is:
\begin{equation}
\small
\begin{aligned}
        \nabla_{o_i} \mathcal{L} &= m_i \cdot p_i - \mathbbm{1}(i=k) \\
        &= 
    \begin{dcases}
    (1 - \alpha \frac{p_{\text{neg}}}{1 - p_{\text{neg}}})p_i - 1, \text{  if } i = k\\
    (1 - \alpha \frac{p_{\text{neg}}}{1 - p_{\text{neg}}})p_i, \quad \quad  \text{if } i \ne k \text{ and } i \neq i_{\text{neg}}\\
    (1 + \alpha)p_i, \quad\quad \quad\quad \quad \text{~~if } i \ne k \text{ and } i = i_{\text{neg}}
    \end{dcases}
    \normalsize
\end{aligned}
\label{eq:ul_analysis}
\end{equation}
where $p_i=[\softmax(\vo)]_i$, $p_{\text{neg}}$ is the probability of the negative-candidate token with index $i_{\text{neg}}$, and $\mathbbm{1}(i=k)$ is the indicator function with $k$ being the index of the ground truth or target token (see the original paper for derivation). 

We know  that as the gradient-based optimization progresses, the gradient norm decreases and converges to near $0$ (\cref{subsec:mle_gradient}-\cref{subsec:method}). To generate a ground truth token, the model must learn to assign the highest probability to it. In other words, the probability assigned by the model to a ground truth token (\ie\ $p_k$) should always increase as the training progresses, or equivalently the norm $|\nabla_{o_k}\mathcal{L}|$ should decrease (monotonic relation). If this is not the case, the model may not learn to predict the ground truth tokens correctly, which in turn hurts the generation quality.

\begin{table*}[ht]
\caption{Results for open-ended generation tasks on the \textbf{Wikitext-103 testset}.
\textbf{ppl}, \textbf{uniq} and \textbf{Rep/l} are computed at BPE-level and the rest are at word-level. The  ``$\uparrow$'' denotes higher value for better performance and ``$\downarrow$'' is the opposite. {Number marked with * are estimated based on the testset.} {The results are averaged over 3 runs with different random seeds. Full results with standard deviation are reported in Appendix~\ref{append:fullresultsopen}.}}
\centering
\scalebox{0.94}{\begin{tabular}  {lccccccccc} 
\toprule
 &  \multicolumn{5}{c}{\textbf{Language Modeling}}
 & \multicolumn{4}{c}{\textbf{Auto Completion}} \\
 \cmidrule(lr){2-6}  \cmidrule(lr){7-10} 

\textbf{Models} & \bf{ppl} $\downarrow$ & \bf{uniq} $\uparrow$ &\bf{Rep/16} $\downarrow$  &\bf{Rep/32} $\downarrow$  & \bf{Rep/128} $\downarrow$ &\bf{Rep-1} $\downarrow$ & \bf{Rep-2} $\downarrow$ & \bf{Rep-3} $\downarrow$  & \bf{uniq-w} $\uparrow$\\
\midrule

MLE             & \bf{13.241} & 12.54k & 0.234 & 0.380 & 0.619 & 0.661 & 0.500 & 0.424  & 16.83k\\
UL ($\alpha=1.0$)    & 16.062 & 13.18k & 0.212 & 0.341 & 0.558 & 0.559 & 0.363 & 0.291 & 19.11k  \\
SG ($\gamma=0.2$)  & 14.203 & \bf{13.61k} & \bf{0.197} & \bf{0.317} & \bf{0.522} & \bf{0.443} & \bf{0.215} & \bf{0.143}  & \bf{22.25k} \\

\midrule
Human                & - & 18.27k & 0.177 & 0.285 & 0.480 & {0.382*} & {0.096*} & {0.037*} & {27.55k*}\\
\bottomrule
\end{tabular}}
\label{table:openended}
\end{table*}

Based on the gradients (Eq. \ref{eq:ul_analysis}), we can identify one case where UL may provide such undesired property. Since the ground truth is always by definition a non-negative token in UL (\ie $i = k \neq i_\text{neg}$), the gradient norm from Eq.~\ref{eq:ul_analysis} is $|\nabla_{o_k} \mathcal{L}| = |\mu_k \cdot p_k - 1|$ where  $\mu_k = (1 - \alpha \frac{p_{\text{neg}}}{1 - p_\text{neg}})$. We see that when $p_\text{neg} > \frac{1}{\alpha+1}$ (\eg when  $\alpha=1$ and $p_{\text{neg}} > 0.5$), $\mu_k$ becomes negative, having the gradient norm $|\nabla_{o_k} \mathcal{L}| = \big|-|\mu_k| \cdot p_k - 1\big| = |\mu_k| \cdot p_k + 1$. In this case, 
the training procedure
will decrease $p_k$  to reduce $|\nabla_{o_k} \mathcal{L}|$, which contradicts with the {optimization principle}. Thus, UL may become less effective in such special cases (subject to the choice of the value of $\alpha$ and $p_{\text{neg}}$). 
Appendix~\ref{append:ulissue} further clarifies this issue using the same notation as the original paper \cite{welleck2019neural}. In contrast, the gradient analysis in Eq.~\ref{eq:sgcases} shows that {ScaleGrad} does not have such
{properties}
in learning to predict ground truth tokens.

\label{sec:method}

\section{Experiments}
\label{sec:exps}

\begin{table*}[ht]
\caption{Results for open-ended generations on \textbf{PTB} testset.
\textbf{ppl}, \textbf{uniq} and \textbf{Rep/l} are computed at BPE-level and the rest are at word-level. The  ``$\uparrow$'' denotes higher value for better performance and ``$\downarrow$'' is the opposite. {Numbers marked with * are estimated based on the testset.}}
\centering
\scalebox{0.94}{\begin{tabular}  {lccccccccc} 
\toprule
 &  \multicolumn{5}{c}{\textbf{Language Modeling}}
 & \multicolumn{4}{c}{\textbf{Auto Completion}} \\
 \cmidrule(lr){2-6}  \cmidrule(lr){7-10} 

\textbf{Models} & \bf{ppl} $\downarrow$ & \bf{uniq} $\uparrow$ &\bf{Rep/16} $\downarrow$  &\bf{Rep/32} $\downarrow$  & \bf{Rep/128} $\downarrow$ &\bf{Rep-1} $\downarrow$ & \bf{Rep-2} $\downarrow$ & \bf{Rep-3} $\downarrow$  & \bf{uniq-w} $\uparrow$\\
\midrule
\bf{PTB}\\
MLE & \bf{33.952} & 5.60k &  0.157  & 0.292  & 0.530  & 0.652 & 0.493 & 0.424 & 6.46k    \\

UL ($\alpha=1.0$)   &  41.232 & 5.96k &  0.139 & 0.260 & 0.476 & 0.533 & 0.333 & 0.259 & 7.60k \\

SG ($\gamma=0.2$) & {40.731} & \bf{6.15k} & \bf{0.126} & \bf{0.231} & \bf{0.426} & \bf{0.417} & \bf{0.198} & \bf{0.131} & \bf{8.42k} \\

\midrule
Human  & - & 8.84k &  0.118 & 0.222 & 0.421 & 0.362* & 0.089* & 0.033* & 11.32k*\\

\bottomrule
\end{tabular}}
\label{table:openendedonptb}
\end{table*}

We showcase the performance of {ScaleGrad} in both open-ended  and directed generation tasks.  
To  verify the effectiveness of our approach, for all the experiments below, we use exactly the same hyper-parameters (except for method-specific ones) and setup as the corresponding baseline unless stated otherwise. All the experimental details, such as model hyper-parameters, training and dataset settings regarding the reproducibility can be found in Appendix \ref{append:hyperparameters}. For qualitative assessments, we show examples of generated texts 
in {Table~\ref{table:example auto-completion in main} and more in} Appendix~\ref{append:examples}. For both open-ended and directed generation tasks, in order to model different {regularization} strengths 
imposed by the methods, 
we choose $\alpha \in \{0.5,1.0,1.5\}$ for unlikelihood training and $\gamma \in \{0.2, 0.5, 0.8\}$ for ScaleGrad.\footnote{$\alpha=1.0$ is recommended by \citet{welleck2019neural}, which can be seen as applying unlikelihood loss with a moderate strength. We use $\alpha=0.5$ and $1.5$ to evaluate for weak and strong strengths.} {The final models are chosen based on their performance on the corresponding development sets.}

\subsection{Open-ended generation}
\label{subsec:opengeneration}
\paragraph{Setup} We consider language modeling and text auto-completion, where we compare the performance of the model trained with {ScaleGrad} against the models trained with MLE and unlikelihood (UL) training introduced lately to mitigate degeneration in open-ended tasks. We follow the same setup as \cite{welleck2019neural}. Specifically, we fine-tune the pre-trained GPT-2 \citep{radford2019language} on Wikitext-103 \citep{merity2016pointer} with a maximum sequence length of 300 tokens. Each model is trained for a maximum of 35k iterations and evaluated based on the perplexity on the validation set after every 1k iterations. We report language modeling results on the testset for each model selected according to the perplexity on the validation set. The same saved models are also used for text  auto-completion, where 50 BPE \citep{sennrich-etal-2016-neural} tokens {(from testset)} are given as prefix and the models are to generate the continuation of 100 next tokens. To evaluate the modeling capability exclusively, following \citet{welleck2019neural}, we apply greedy decoding in all our experiments in this section. Later, in \cref{subsec:analysis on openended}, we analyze how our method performs
with different decoding methods.

{In language modeling, we measure the generation quality by the standard perplexity ({ppl}), {and {Rep/$l$} and  {`uniq'} 
measures of token-level distribution}
as \cite{welleck2019neural}.} {Rep/$l$} measures the number of times that a token from the previous $l$ tokens is repeated, when generating the following token; in our case, $l \in \{16,32,128\}$. The {`uniq'} is defined as the number of unique next-token predictions on a test/validation set. For auto-completion, we report the repetition ratios of n-gram words ({Rep-n})  as well as the number of unique words ({uniq-w}) that are used during generation on the testset.

\paragraph{Results}

We present our main results on the \emph{testset} in Table~\ref{table:openended}. The results with different hyper-parameters for both methods on the \emph{validation} set are reported in Appendix~\ref{append:fullresultsopen} and in \cref{subsec:hyper}.
From Table~\ref{table:openended}, we notice that in language modeling, the model trained with {ScaleGrad} (SG) yields a token distribution that is much closer to human, while maintaining a lower perplexity. {Compared to the 
{UL} baseline, SG achieves 1\%, 2\%, 4\% lower repetitions in Rep/16, Rep/32 and Rep/128, respectively, while having 11\% lower perplexity. It also uses more \textbf{uniq}ue tokens compared to others (\eg 3\% more compared to UL training)}. Overall, our method significantly improves the token-level distribution and keeps a high generation quality. 
In auto-completion, from the quantitative perspective,  SG produces texts with much fewer repetitive n-grams compared to  MLE and UL. It uses nearly 5.5k more unique words compared to the MLE baseline, which {agrees with} the purpose of making the model learn to use novel tokens in training. 

\paragraph{Human evaluation} We have conducted a user study to verify the quality of generated texts. 
The study is conducted for two pairs of systems ({SG} vs. UL, {SG} vs. MLE). For each pair, we randomly choose the same 100 prefixes for the systems to produce their own continuations and 
ask two native speakers of English to judge which text is the better continuation of the given prefix in terms of their \emph{relevance} to the prefix, \emph{grammaticality} and \emph{readability}. More details about the study can be found in Appendix~\ref{append:human}.

\begin{table}[t!]
     \centering
    \captionof{table}{Human evaluation results for auto-completion. \textbf{\% Agr.} is the percentage agreement and \textbf{AC1} denotes Gwet's AC1/gamma coefficient. Winners are marked in \textbf{bold}.}
   \label{table:humanevaluation}
    \scalebox{0.94}{\begin{tabular}{lccc}
    \toprule
       & \bf{Win Rate} & \bf{\% Agr.} & \bf{AC1}  \\
    \midrule
    \textbf{SG} vs MLE & 84.0\% & 84.0\% & 0.78 \\
    \textbf{SG} vs UL & 70.5\% & 79.0\% & 0.64 \\
    \bottomrule
   \end{tabular}}
\end{table}

From the results in Table~\ref{table:humanevaluation}, we can observe that the texts produced by the models trained with {ScaleGrad (SG)} are preferred by the human users in most of the cases, \ie 84.0\% and 70.5\%, respectively. We also compute the percentage agreement and 
{chance-correlated Gwet's AC1/gamma coefficient \citep{gwetac1}  as inter-user agreement to verify the reliability of the study (details in Appendix~\ref{append:human}). We see that the agreements are substantial in both measures.}

\paragraph{Generalizability}
\label{append:ongeneralizability}
To further verify the generalizability (\ie different datasets and domains) of our method, apart from WikiText-103 \citep{merity2016pointer}, we evaluate the models on two other datasets: Penn TreeBank or PTB \citep{marcus-etal-1993-building} and IMDB \citep{maas-EtAl:2011:ACL-HLT2011}. In particular, after fine-tuning GPT-2 with different training strategies (MLE, SG and UL) on WikiText-103 training data, we test the language modeling and auto-completion performance on the other two datasets. For PTB, we use the standard testset. As for IMDB, we randomly sample 500 movie reviews from the dataset.

In Table~\ref{table:openendedonptb}, we show the experimental results on the PTB testset, from which we can see that SG consistently improves over the MLE baseline in degeneration while possessing an acceptable increase in perplexity, and it outperforms UL consistently. Additionally, we present the results on IMDB movies review in Table~\ref{table:openendedonimdb} in Appendix~\ref{append:imdb}, where we observe similar performance trending as in the experiment on PTB testset. From the two experiments, we can draw the conclusion that our method, SG, is capable of generalizing well to different datasets and domains. 
{Examples of generated text for auto completion task can be found in Appendix~\ref{append:examples}.}

\subsection{Directed generation}
\label{subsec:directgen}

For directed generation, we consider two tasks: image paragraph captioning and text summarization. 

\begin{table}[ht]
\centering
    \captionof{table}{Results for image paragraph captioning on the Visual Genome testset.}
    \label{table:resultsimg-para}
    \scalebox{0.90}{\begin{tabular}{lc}
    \toprule
    \bf{Models} &  \bf{CIDEr}  \\
    \midrule
    \text{MLE} w/o 3-block & {10.51}\\
    UL w/o 3-block ($\alpha$=0.5)  \quad \quad \quad& 14.65 \\
    {SG} w/o 3-block ($\gamma$=0.5) & \bf{19.42} \\
    \midrule
    \text{MLE} w/ 3-block & 22.77\\
    UL w/ 3-block ($\alpha$=0.5) & {22.25} \\
    {SG} w/ 3-block ($\gamma$=0.5) & \bf{24.62} \\
    \bottomrule
   \end{tabular}}
  \vspace{-1em}
\end{table}
\subsubsection{Image paragraph captioning}
\paragraph{Setup} We use the captioning model proposed by  \citet{melas-kyriazi-etal-2018-training} as the baseline, which comprises a CNN encoder that is pre-trained for object detection and a 1-layer LSTM decoder. The models are trained and evaluated on the paragraph captioning dataset, Visual Genome  \citep{krause2016paragraphs}. We train the model with SG and compare it to the ones trained with MLE and UL. The performance is measured by CIDEr \citep{vedantam2015cider}, which computes TF-IDF weighted n-gram overlaps between {the model generated captions and the reference captions.}  {We follow \citet{melas-kyriazi-etal-2018-training} to apply greedy inference since beam search did not yield any further gain.}

\vspace{-1em}
\paragraph{Results} 
Table~\ref{table:resultsimg-para} {shows the CIDEr scores for different training methods on Visual Genome testset with and without tri-gram blocking \citep{Paulus2018ICLR} during inference.} Without tri-gram blocking, MLE produces texts that are full of repetitive phrases (see \Cref{append:examples} for  examples), which leads to a low CIDEr score. When UL or {SG} is incorporated, the performance has been notably improved from {10.51} to 14.65 and 19.42, respectively.
When tri-gram blocking is applied, our method is still capable of yielding 1.85 point improvement. This is because {SG} further improves the token-level degeneration {on top of tri-gram blocking.} In contrast, the model trained with UL has a slightly worse CIDEr score compared to the MLE baseline. {We analyze n-gram level degeneration further in \cref{subsec:analysis on directed}.}

\begin{table}[ht]
\caption{Experimental results for text summarization on CNN/DM and NYT50 testsets. \textbf{R-1}, \textbf{R-2} and \textbf{R-L} stand for F1-based ROUGE-1, ROUGE-2 and ROUGE-L. \textbf{WMD-1} denotes 1-gram MoverScore.}
\centering
\scalebox{0.81}{\begin{tabular}{lcccc} 
\toprule
\bf{Models} & \bf{R-1} & \bf{R-2} & \bf{R-L} & \bf{WMD-1}  \\
\midrule
\textbf{CNN/DM} \\
BertSum w/ MLE & 41.87 & 19.42 & 38.93 & 19.89 \\
BertSum w/ UL ($\alpha=0.5$) \quad \ & 42.03 & 19.36 & 39.09 &  20.21 \\
BertSum w/ SG ($\gamma$ = 0.8) & \bf{42.19} & \bf{19.53} & \bf{39.25} & \bf{20.23} \\
\midrule
\textbf{NYT50}\\
BertSum w/ MLE & 48.73 & 31.00 & 45.23 & 28.73 \\
BertSum w/ UL ($\alpha=0.5$) & {48.54} & {30.73} & {44.99} & {28.50} \\
BertSum w/ SG ($\gamma$ = 0.8) & \bf{49.29} & \bf{31.30} & \bf{45.78} & \bf{29.14} \\
\bottomrule
\end{tabular}}
\label{table:summarization}
\end{table}

\begin{table*}[t]
    \centering
    \caption{An example of generated continuations for the same prefix.}
    \label{table:example auto-completion in main}
    \scalebox{0.90}{\begin{tabular}{p{1.07\textwidth}}
        \toprule
        \textbf{Prefix}: You's patriotism, and Mei's reflections on the quotidian are a few examples. More broadly, Du Fu's work in transforming the lushi from mere word play into ``a vehicle for serious poetic utterance" \\
        \midrule
         \textbf{MLE}:
         is a major contribution to the genre.  The first of the three major works of the genre,  was written in the early 1920s, and was the first of the three to be published in English. It was published in the United States in the United States, and was translated into French in the 1930s. The work was the first of the three \\
         \midrule
         \textbf{UL}:
         is a major influence on the modern Chinese novel. The modern Chinese novel is often characterized as being ``literary in nature", but it is not limited to literary genres. The modern Chinese novel has been described as ``a literary genre that has been influenced by the Western literary tradition". The modern Chinese novel has been described as ``a literary genre that has been influenced by the Western literary tradition" \\
         \midrule
         \textbf{SG}:
         is a powerful example of how the Chinese literary tradition can be used to explore themes of nationalism and self determination. In addition to his own poetry, Du Fu wrote several works of prose. The first was a collection of short stories called ``The Story of the Three Sisters", which was published in 1885. It was followed by a series of novels, including the first two volumes of the Song dynasty epic poem ``The Three Sisters" \\
         \bottomrule
    \end{tabular}}
\vspace{-0.5em}
\end{table*}

\subsubsection{Abstractive text summarization}

\paragraph{Setup} We use the abstractive summarization model BertSum \citep{liu-lapata-2019-text} as our baseline, which  adopts a Transformer architecture to take advantage of pre-trained BERT \citep{devlin-etal-2019-bert} as the encoder. {At the first stage,} the encoder is trained with an extractive summarization objective (binary classification for sentence selection). At the second stage, it initializes the decoder randomly and (re)trains the entire  encoder-decoder model with an abstractive (or generative) objective. 
{For our experiments, we take the encoder that was trained at the first stage and train the entire (abstractive) model with different training methods (MLE, UL and SG)}
using the default training setup on two benchmark datasets: CNN/DM \citep{hermann2015CNN,nallapati-etal-2016-abstractive} and NYT50 \citep{durrett-etal-2016-learning}. 
During inference, length normalization \citep{google-nmt}, tri-gram blocking {and beam search (beam size = 5)} are used as in \citep{liu-lapata-2019-text}.

We evaluate performance of the models with the standard {F1-based}  ROUGE \citep{lin2004rouge} scores  (R-1, R-2, R-L) and a model-based evaluation MoverScore \citep{zhao-etal-2019-moverscore}, which computes the Word Mover Distance (WMD) between the reference summary and generated summary based on the {representations from BERT.} We report 1-gram MoverScore ({WMD-1}), which has been proven to have higher correlation with human than other metrics  \citep{zhao-etal-2019-moverscore}.

\vspace{-1em}
\paragraph{Results} From Table~\ref{table:summarization}, we notice that on CNN/DM, the model trained with SG  outperforms the models trained with MLE and UL when measured by ROUGE. In WMD-1, UL yields similar performance as ours. Both SG and UL further improve over the  MLE baseline. The improvements imply that token-level degeneration may still exist even when tri-gram blocking is applied. On NYT-50, UL underperforms MLE, while our method improves in all measures. In \cref{subsec:issueoful}, we discussed a possible reason behind UL's  underperformance from a gradient perspective.

\section{Analysis {of ScaleGrad}}
\label{sec:analysis}

\begin{table}[ht]
\centering
\caption{Results of different decoding strategies with ScaleGrad training for auto-completion on WikiText-103 testset.} 
\centering
\label{table:decodingmethods}
\scalebox{0.84}{\begin{tabular} {lcccc}
\toprule
\bf{Approaches} & \bf{Rep-1} & \bf{Rep-2} & \bf{Rep-3} & \bf{uniq-w} \\ 
\midrule
{SG+}Greedy Search & {0.443} & {0.215} & {0.143}  & {22.25k}\\
SG+Beam Search ($b$ = 6) &  0.453 & 0.250 & 0.171 & 8.32k\\
SG+Top-$p$ ($p$ = 0.3) &  0.356 & 0.107 & 0.049 & 30.48k\\

SG+Top-$k$ ($k$ = 40) &  0.254 & 0.039 & 0.012 & 39.50k\\
\bottomrule
\end{tabular}}
\end{table}

{After comparing with UL and MLE on both directed and open-ended generation tasks, we now analyze ScaleGrad from different perspectives to gain more insights.}

\subsection{Open-ended generation}
\label{subsec:analysis on openended}
\paragraph{Compatibility with decoding strategies} 

One advantage of SG training is that it is compatible with {{decoding-based methods}}. {One can choose different decoding strategies based on the specific needs.} Table~\ref{table:decodingmethods}  provides the results of different decoding strategies used along with {SG training} for text auto-completion (results for other variations and baselines are in  Appendix~\ref{append:decodingstrategy}). We observe that beam search, even with larger beam size, is not effective in mitigating the degeneration issue, which accords with the observation in \citep{holtzman2019curious}. As expected, stochastic decoding, top-$k$ and nucleus  (top-$p$) sampling, help to further reduce repetition. This sets good examples of combining training and decoding strategies for the task in hand.

\begin{figure}[ht]
    \centering
    \includegraphics[width=0.48\textwidth]{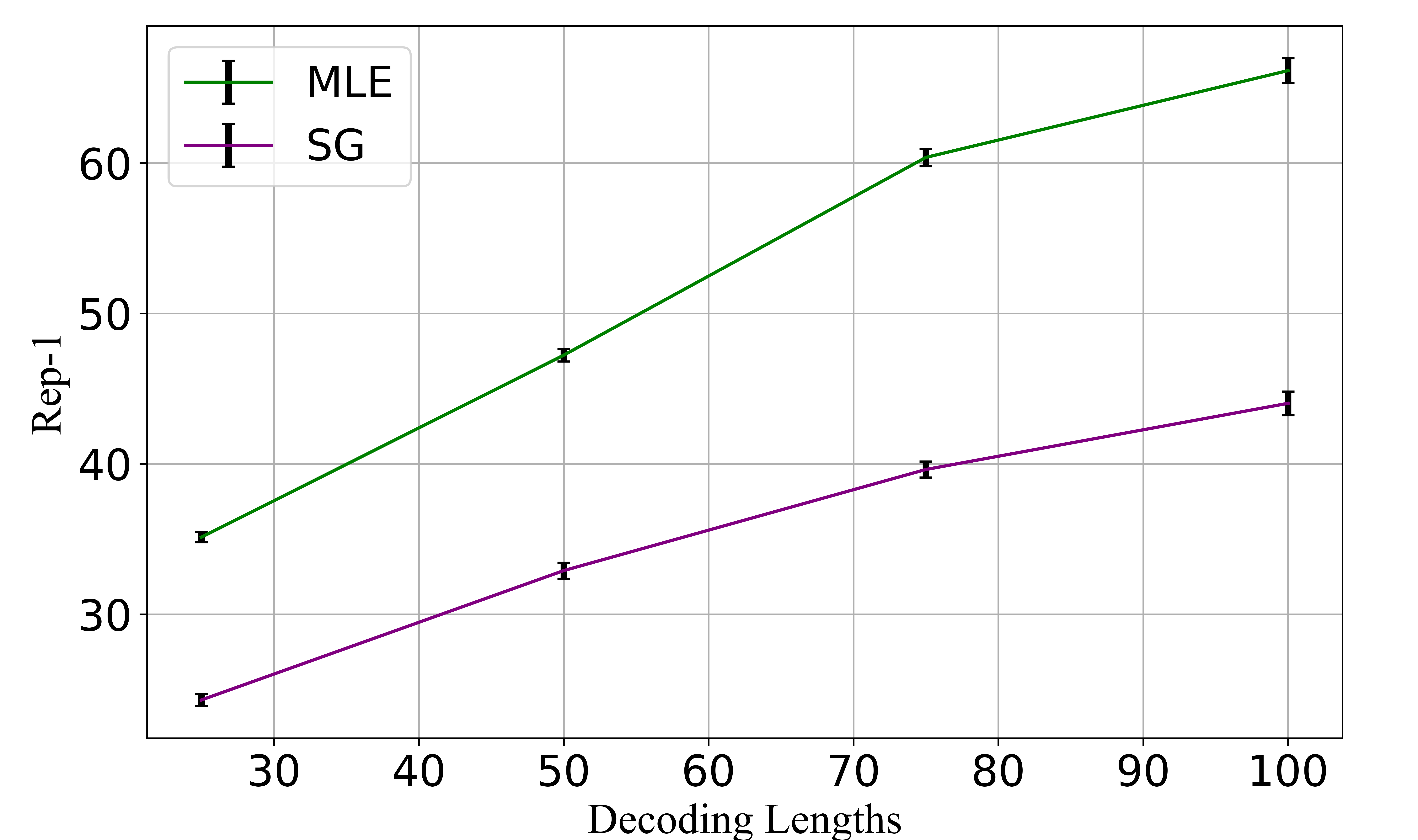}
    \caption{\textbf{Rep-1} in auto completion with different decoding lengths. All the numbers are computed based on the results from 3 runs with different random seeds.}
    \label{fig:lengths}
    \vspace{-1em}
\end{figure}

\paragraph{Auto completion with different decoding lengths}
\label{append:generalizability lenghts}
From a practice point of view, we analyze how SG 
performs in text generation with varied decoding lengths.
In Figure~\ref{fig:lengths}, we show the \textbf{Rep-1} of generated text from the auto completion task with the constraints in different decoding (continuation) lengths. We see that compared to MLE counterpart, SG yields consistent improvements on \textbf{Rep-1}, or token-level degeneration, regardless the different decoding lengths, which further verifies the effectiveness and generalizability of our method.

\begin{figure*}[ht]
     \centering
     \begin{subfigure}[b]{0.32\textwidth}
         \centering
         \includegraphics[width=\textwidth]{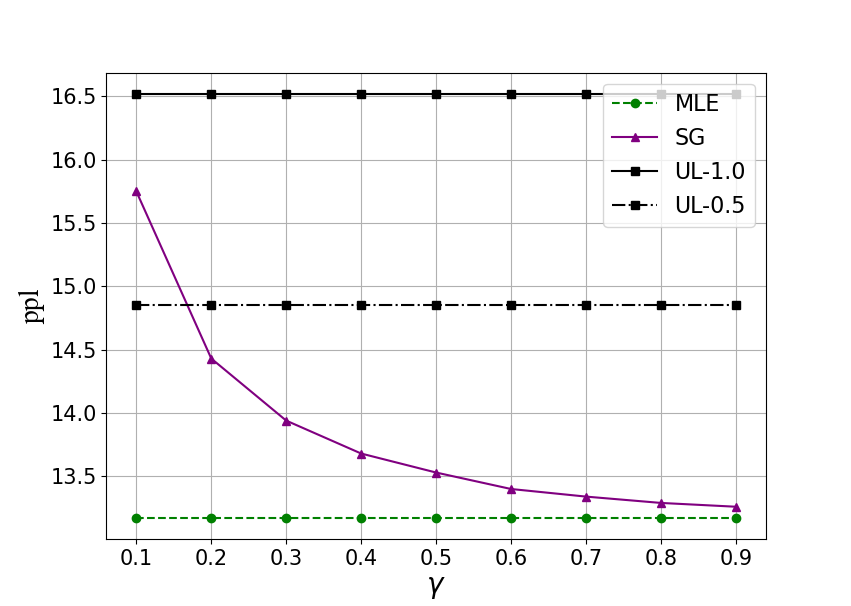}
         \caption{Perplexity}
         \label{fig:ppl}
     \end{subfigure}
     \hfill
     \begin{subfigure}[b]{0.32\textwidth}
         \centering
         \includegraphics[width=\textwidth]{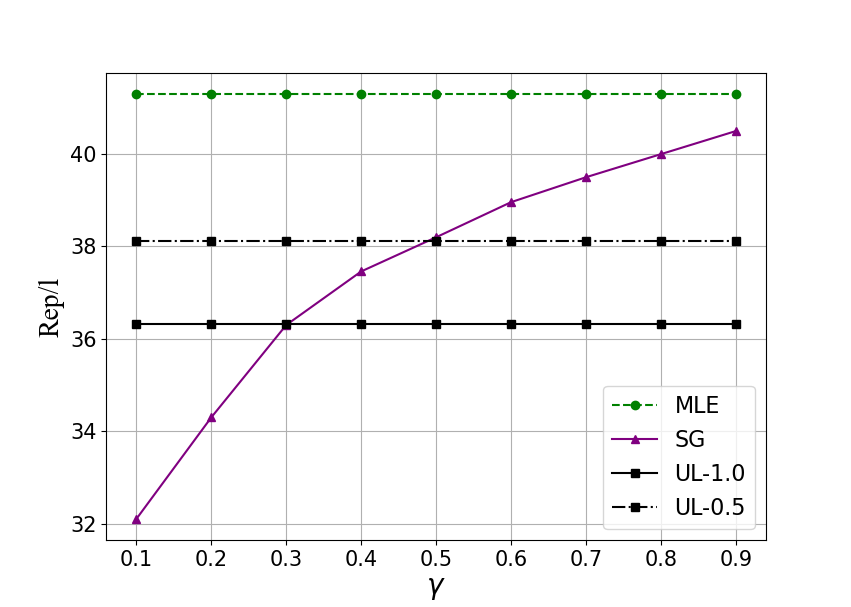}
         \caption{Rep/$l$}
         \label{fig:repl}
     \end{subfigure}
     \hfill
     \begin{subfigure}[b]{0.32\textwidth}
         \centering
         \includegraphics[width=\textwidth]{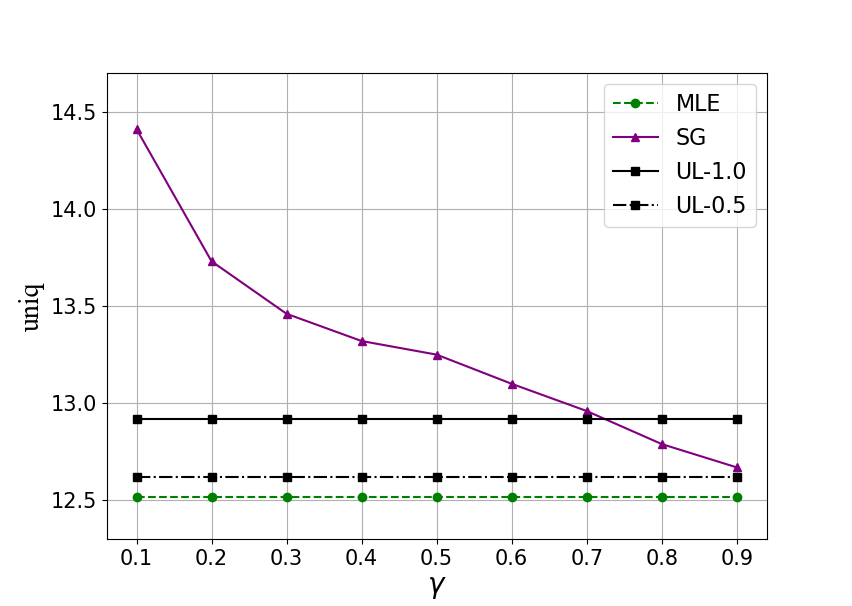}
         \caption{\# of unique tokens}
         \label{fig:uniq}
     \end{subfigure}
        \caption{Hyper-parameter ($\gamma$) sensitivity in the language modeling task on \textbf{Wikitext-103 development} set (best viewed in color). Rep/$l$ is computed as the average of Rep/$16$, Rep/$32$ and Rep/$128$. UL-1.0 and UL-0.5 represent unlikelihood training with $\alpha$=1.0 and 0.5, respectively. $\alpha = 1.5$ is not included as it incurs significantly higher perplexity compared to others. 
        Individual Rep/$l$ results can be found in Appendix~\ref{append:hyper-sensi}.}
        \label{fig:three graphs}
\end{figure*}

\subsection{Directed generation}
\label{subsec:analysis on directed}
 \paragraph{Comparison with stochastic decoding} 
 Although top-$p$ and top-$k$ sampling have been proven successful in open-ended generation, they have not been tested in directed generation tasks. In order to see if they could lead to the same improvements as SG training, we conduct additional experiments with the BertSum summmarization model, whose underlying language model is more mature due to the involvement of BERT, compared to the image paragraph captioning model. {For the interested readers, we also provide the results of stochastic decoding on image paragraph captioning in Appendix~\ref{append:stochastic for img}.}

Table~\ref{table:toppkonsumm} shows the performance for stochastic decoding in BertSum trained with MLE. Since ROUGE-1 measures the exact 1-gram overlaps between reference and generated summaries, it may not be sufficient to evaluate the performance of stochastic decoding methods, which may generate more diverse output while conveying the same meaning. Therefore, we also report the MoverScore that is capable of considering the semantic similarity rather than just n-gram overlaps. Both the ROUGE and MoverScore in Table~\ref{table:toppkonsumm} lead to the conclusion that stochastic decoding methods significantly lower the performance compared to the standard beam search. This implies that they may not be a good fit for directed generation tasks. In contrast, SG possesses a wider applicability in mitigating degeneration issues as shown earlier in \Cref{table:summarization}.
 
\begin{table}[ht]
    \centering
    \captionof{table}{Summarization results (F1-based ROUGE-1 and MoverScore) for stochastic decoding on NYT50 testset.}
   \label{table:toppkonsumm}
    \scalebox{1}{\begin{tabular}{lcc}
    \toprule
    \bf{Models} & \bf{ROUGE-1} & \bf{WMD-1}\\
    \midrule
    Top-$p$ (p=0.3) & 45.44 & 24.61 \\
    Top-$p$ (p=0.9) & 42.33 &  21.67 \\
    Top-$k$ (k=40) & 41.23 & 20.70\\
    Top-$k$ (k=100) \ \quad \quad \ & 40.86 & 20.38\\
    \midrule
    Baseline & 48.73 & 28.73 \\
    \bottomrule
    \end{tabular}}
    \vspace{-1em}
\end{table}

\paragraph{N-gram degeneration} 

To investigate further how SG minimizes degeneration and helps to improve the performance in automatic measures, we compute the n-gram repetition ratios of the outputs from the image captioning model \citep{melas-kyriazi-etal-2018-training} and report the numbers in Table~\ref{table:degeneration analyisis}.\footnote{Since \citet{melas-kyriazi-etal-2018-training} used a soft tri-gram blocking, some of the duplicate tri-grams still remain.} 
Compared to human, the MLE baseline has significantly higher repetitions, thus having the lowest CIDEr score (\Cref{table:resultsimg-para}). With SG, the model 
yields a much better repetition ratio, which explains the notable performance boost in CIDEr. Tri-gram blocking resolves the issue of 3- or higher n-gram degeneration in a hard-coded way, improving CIDEr significantly. However, the token and 2-gram repetitions still remain high and improvable in MLE with tri-gram blocking. When both tri-gram blocking and SG are applied, the generated texts have the lowest and most human-like repetitions. 
\begin{table}[ht]
 \centering
    \captionof{table}{Degeneration analysis for image paragraph captioning with/without tri-gram blocking. Numbers in bold are closest to human.} 
   \label{table:degeneration analyisis}
   \scalebox{1}{\begin{tabular}{lccc}
    \toprule
    \bf{Models} & \bf{Rep-1} & \bf{Rep-2} & \bf{Rep-3}\\
    \midrule
    MLE & 0.723 & 0.587 & 0.530\\
    {SG}  & {0.500} & {0.270} & 0.195\\
    MLE w/ 3-block \ \quad \quad \ & 0.575 & {0.271} & {0.094}\\
    {SG} w/ 3-block & \bf{0.440} & \bf{0.146} & \bf{0.037} \\
    \midrule
    Human & 0.421 & 0.123 & 0.042 \\
    \bottomrule
    \end{tabular}}
    \vspace{-1em}
\end{table}

\subsection{Hyper-parameter sensitivity} \label{subsec:hyper}
Towards better usage and understanding of ScaleGrad, we show how the key metrics in language modeling  
change with the hyper-parameter $\gamma$ in Figure~\ref{fig:three graphs}.\footnote{Note that for our main results in \cref{sec:exps}, we only search hyper-parameters from 3 chosen values. More numbers of $\gamma$ in Figure~\ref{fig:three graphs} is intended to show the hyper-parameter sensitivity of ScaleGrad. One should not regard this as unfair comparison where different numbers of hyper-parameter are explored for different methods.} As discussed, a smaller value of $\gamma$ incurs a stronger push to use novel tokens, giving higher perplexity and more unique tokens. {Having a low perplexity and a low repetition ratio could be seen as a trade-off between general generation quality and diversity. However, we observe that when UL achieves similar performance in Rep/$l$ with SG, \ie when $\gamma=0.5$, $\alpha=0.5$ and $\gamma = 0.3$, $\alpha=1.0$ (\cref{fig:repl}), it exhibits much higher perplexity compared to SG with a difference of 1.35 and 2.58, respectively (\cref{fig:ppl}). Similarly, when both methods have similar performance on perplexity, \ie when $\gamma=0.2$ and $\alpha=0.5$ (\cref{fig:ppl}), SG yields 3.82\% lower in Rep/$l$ (\cref{fig:repl}) and uses 1.11k more unique tokens (\cref{fig:uniq}). In summary, SG is able to reduce the degeneration without detracting much from the generation quality.}

 In general, $\gamma$ in ScaleGrad can be chosen based on the performance of the baseline model. If the baseline produces many repetitive tokens/phrases (\eg image paragraph captioning experiments), a smaller value of $\gamma$ should be used. Conversely, in tasks with less degeneration (\eg summarization experiments), a larger $\gamma$ can be used to further improve the unigram and bigram level degeneration without affecting the perplexity much.

\section{Conclusion}
We have introduced a novel training method, called {ScaleGrad}, directly modifying the gradient of the standard MLE objective to remedy the text degeneration issues. 
The improvement verified by both automatic metrics and human evaluation against the baselines in extensive experiments across different tasks in open-ended and directed generation and different architectures (\ie LSTM and Transformer) demonstrate the effectiveness and generalizability of our method. Further analysis shows that
{ScaleGrad} yields token distributions that are much closer to human-written texts compared to the baselines. Our method brings a good alternative to current training strategies for language generation.

In future, we plan to extend the idea in two directions. First, we would like to repurpose the definition of the set of the tokens that ScaleGrad operates on (\ie the novel token set) to enable the model to realize other objectives, \eg \citet{ying-2021-endpoint} has successfully adapted ScaleGrad to prevent early endpointing for online automatic speech recognition.
Second, we would like to investigate a  mechanism to dynamically adjust the hyper-parameter $\gamma$ in  the decoding steps such that the model could learn with different degrees of strength depending on the context.

\bibliography{ref}
\bibliographystyle{icml2021}

\clearpage
\appendix
\section{Derivations}
\label{append:derivations}

\paragraph{Derivation of the gradient of loss \wrt logit} 
We follow the same notation as in the main paper. At time step $t$, assuming that the pre-softmax scores (\ie\ logits) are denoted as $\vo^t$ over the vocabulary $\sV$, where $o_i^t$ denotes the score for the token with index $i$ in the vocabulary. Similarly, we have $p_i^t = [\softmax(\vo^t)]_i$. Let $k$ denote the index of the  ground truth token at step $t$.

The cross entropy loss at step $t$ is given as  (we omit $t$ for notational simplicity):
\begin{equation}
    \mathcal{L} = - \sum_i y_i \log p_i
\end{equation}

where $y_i=1$ if $i=k$, otherwise $y_i = 0$. Thus the
loss function can be rewritten as:
\begin{equation}
    \mathcal{L} = - \log p_k = -  \log(\frac{e^{o_k}}{\sum_j e^{o_j}}) =  \log(\sum_j e^{o_j}) - o_k 
\end{equation}

Therefore, we can derive the partial derivative of the loss \wrt the logit $o_i$ as follows.

\begin{equation}
\begin{split}
    \nabla_{o_i} \mathcal{L} & =  \nabla_{o_i} \log(\sum_j e^{o_j}) - \nabla_{o_i}o_k \\
     &  =  \frac{1}{\sum_je^{o_j}} \cdot \nabla_{o_i} (\sum_je^{o_j})                - \mathbbm{1}(i = k)  \\
     & = \frac{e^{o_i}}{\sum_je^{o_j}} - \mathbbm{1}(i = k) \\
     & = p_i - \mathbbm{1}(i = k)
\end{split}
\end{equation}

\paragraph{Derivation of the gradient of ScaleGrad \wrt logit}
We first denote $\sS_{\text{novel}}$ as the novel token set at current time step and $\sV' = \sV \setminus \sS_{\text{novel}}$.

Suppose the current target token belongs to the novel token set, \ie $k \in \sS_{\text{novel}}$. The scaling equation for target token can be rewritten into the function of logits as follows.
\begin{equation}
\begin{aligned}
    \tilde{p}_k &= \frac{\gamma \cdot p_k}{\gamma \sum_{j \in \sS_{\text{novel}}} p_j + \sum_{j \in \sV'} p_j} \\
    &= \frac{\gamma \cdot \frac{e^{o_k}}{\sum_{m \in \sV }e^{o_m}}}{ \gamma \sum_{j \in \sS_{\text{novel}}} \frac{e^{o_j}}{ \sum_{m \in \sV }e^{o_m}  }       +     \sum_{j \in \sV'} \frac{e^{o_j}}{\sum_{m \in \sV }e^{o_m}}   } \\
    &= \frac{ \gamma \cdot e^{o_k}  }{ \gamma \sum_{j \in \sS_{\text{novel}}} e^{o_j} + \sum_{j \in \sV'} e^{o_{j}}      }
\end{aligned}
\end{equation}

For the notational simplicity, we notate $a = (\gamma \sum_{j \in \sS_{\text{novel}}} e^{o_j} + \sum_{j \in \sV'} e^{o_{j}} )  $. 
The loss function can be rewritten accordingly as:
\begin{equation}
\begin{aligned}
\mathcal{L} &= - \log(\tilde{p}_k) \\
&= - \log \frac{ \gamma \cdot e^{o_k}}{a} = \log a - \log (\gamma \cdot e^{o_k})\\
\end{aligned}
\end{equation}

We thus have the gradient of the SG loss \wrt the logit ($o_i$) as follows:
\begin{equation}
\begin{aligned}
\nabla_{o_i} \mathcal{L} &= \nabla_{o_i} \log a -  \nabla_{o_i} \log (\gamma \cdot e^{o_k})\\
&= \frac{1}{a} \cdot \nabla_{o_i} a - \frac{1}{\gamma \cdot e^{o_k}} \cdot \nabla_{o_i} (\gamma \cdot e^{o_k}) \\
&= \frac{1}{a} \cdot (\gamma \cdot e^{o_i} \mathbbm{1}(i \in \sS_{\text{novel}}) + e^{o_i} \mathbbm{1}(i \in \sV') ) \\
& \quad - \mathbbm{1}(i=k)\\
&= \begin{dcases}
 \frac{\gamma \cdot e^{o_k}}{a} -1, \quad \text{ if } i = k \text{ and } i \in \sS_{\text{novel}} \\
 \frac{\gamma \cdot e^{o_i}}{a} , \quad \text{ if } i \neq k \text{ and } i \in \sS_{\text{novel}} \\
 \frac{e^{o_k}}{a} - 1, \quad \text{ if } i = k \text{ and } i \notin \sS_{\text{novel}} \\
 \frac{e^{o_i}}{a}, \quad \text{ if } i \neq k \text{ and } i \notin \sS_{\text{novel}} \\
\end{dcases}\\
&= \begin{dcases}
        \lambda_i \cdot {p}_k - 1,  \quad \text{ if } i = k \text{ and } i \in \sS_{\text{novel}}  \\ 
        \lambda_i \cdot {p}_i, \quad \text{ if } i \neq k \text{ and } i \in \sS_{\text{novel}} \\
        \alpha_i \cdot {p}_k - 1, \quad \text{ if } i = k \text{ and } i \notin \sS_{\text{novel}} \\
        \alpha_i \cdot {p}_i, \quad \text{ if } i \neq k \text{ and } i \notin \sS_{\text{novel}}\\
        \end{dcases}
\end{aligned}
\end{equation}

Similarly, it is easy to derive the same results when current target token does not belong to the novel token set.

\section{Novel token set illustration}
\label{append:novelsetfigure}
\begin{figure*}[ht]
    \centering
    \includegraphics[width=0.8\textwidth]{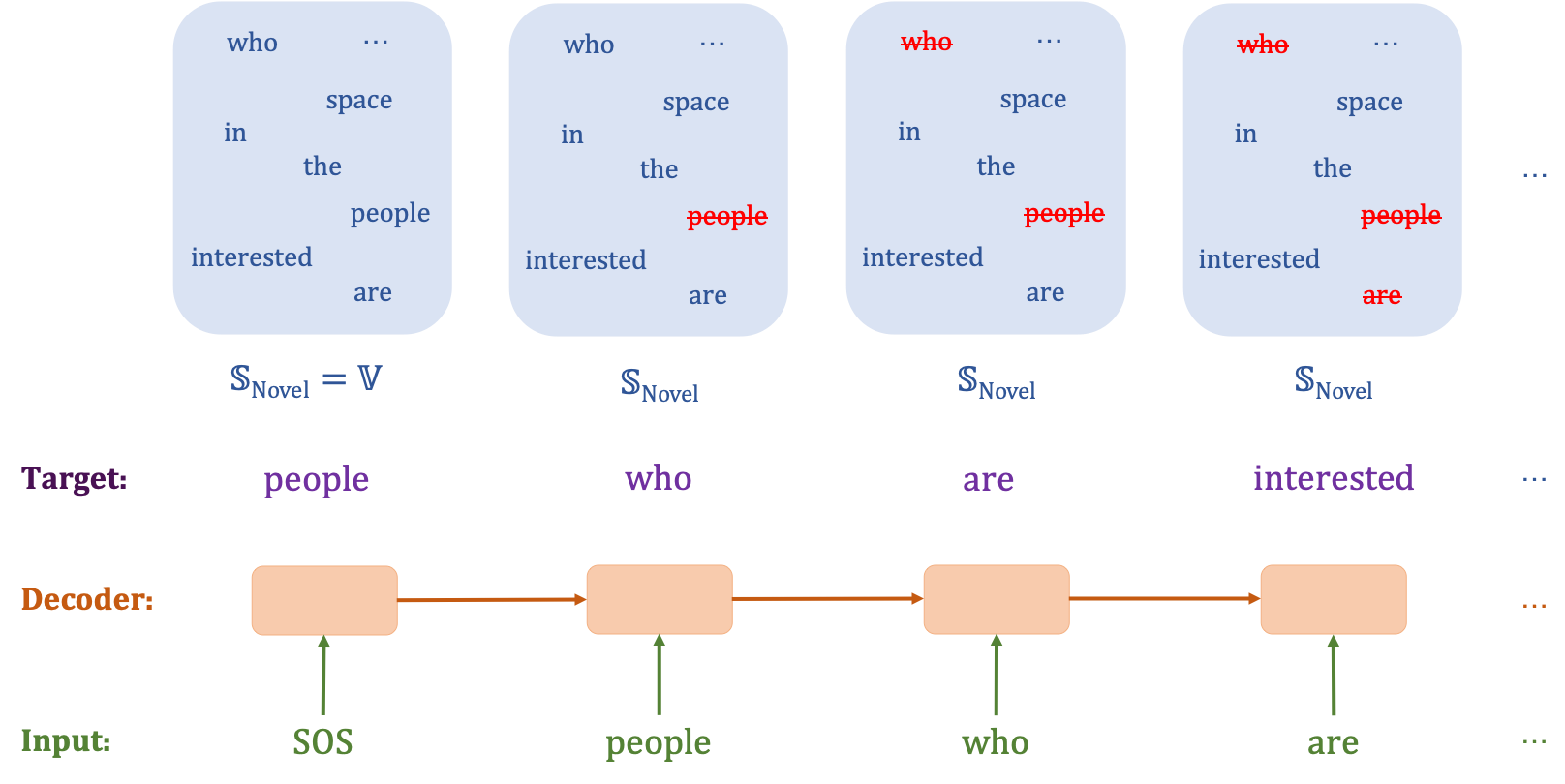}
    \caption{An illustration of how the novel token set changes as decoding proceeds for the sentence ``people who are interested ...". The words marked in purple are the target words that the model is learning to predict at each decoding step. }
    \label{fig:novelset}
\end{figure*}

Figure~\ref{fig:novelset} shows an example of how the novel token set changes when the model is learning to predict the sentence ``people who are interested ..". At beginning, the novel token set $\sS_{\text{novel}}$ is equivalent to the vocabulary $\sV$. The size of the novel token set shrinks as the decoding proceeds.

\section{{Undesired property of UL training}} \label{append:ulissue}

We use the same notation as \citet{welleck2019neural} to explain the undesired UL property. From their paper (page 4): 

With a single negative candidate, the (negative) gradient is:

\begin{equation}
\small
\begin{aligned}
        \nabla \mathcal{L}_a &= x^* - m \odot p, \\  
        &\text{where } m = \begin{dcases} (1 - \alpha \frac{p_{\text{neg}}}{1 - p_{\text{neg}}}) \quad \quad \text{if } i \neq i_{\text{neg}} \\
        (1 + \alpha) \quad \quad \quad \quad \quad \text{ ~if }  i = i_{\text{neg}}
        \end{dcases} \\
    \normalsize
\end{aligned}
\label{eq:ul_grad}
\end{equation}
where $x^{*} \in \{ 0, 1 \}^{\gV}$ is a one-hot ground-truth vector, $m \in \R^{\gV}$, $p = p_{\theta}(\cdot|x_{< t})$, and $p_{\text{neg}}$ is the
probability of the negative candidate at index $i_\text{neg}$.

As the paper says (page 5):

``.... At the ground-truth token index $i^{*}$, the unlikelihood gradient is positive, increasing the ground-truth token’s
probability with a magnitude that grows with $p_{\text{neg}}$.  Conversely, at the negative candidate index $i_\text{neg}$ the gradient is negative. At all other token indices $i \notin \{i^{*}, i_\text{neg} \}$, the gradient moves from negative
to positive as $p_{\text{neg}}$ increases. For instance, with $\alpha = 1.0$ the gradient increases the probability of
each token $x_i$ when the model assigns high probability to the negative candidate ($p_{\text{neg}} > 0.5$). ''

{We notice that at the ground-truth token index $i^{*}$, with $\alpha = 1.0$ and $p_{\text{neg}} > 0.5$, the gradient norm is  $|\nabla \mathcal{L}_a| = 1 + |m| \cdot p^{*}$. The model will therefore decrease $p^{*}$ to reduce $|\nabla \mathcal{L}_a|$, which is against our optimization principle.}

\section{Human evaluation details}
\label{append:human}
We conduct the human evaluation for two pairs of systems \ie {SG} vs. MLE and {SG} vs. UL. {For each pair, the models generate their own continuations based on the same 100 randomly chosen prefixes.  Two native speakers of English are then asked} to evaluate the generated texts independently. During the study, users are instructed to judge which generated text is a better continuation of the prefix based on the overall quality (\eg readability, relevance to the prefix, grammar, and fluency). 

\begin{figure*}[ht]
    \centering
    \includegraphics[width=0.98\textwidth]{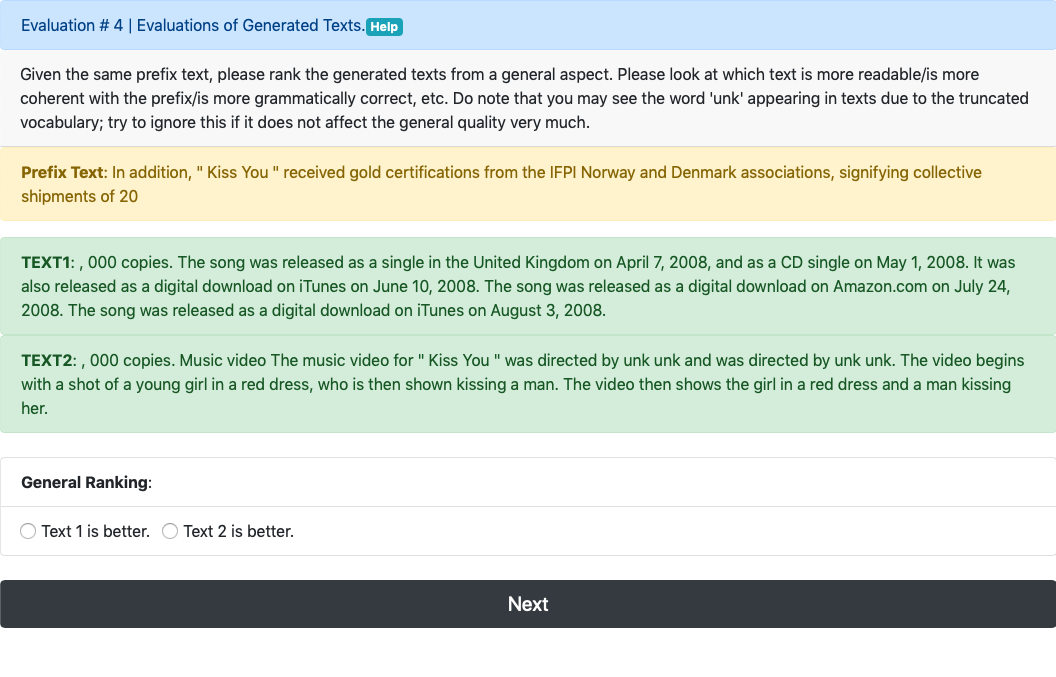}
    \caption{Human evaluation interface}
    \label{fig:interface}
\end{figure*}

The \textbf{Win Rate} in Table~\ref{table:humanevaluation} is calculated as the total number of times that two users prefer the texts produced by the winner divided by the total number of cases in the evaluation ($2 \times 100 = 200$). To get a reliable human study, we also compute the percentage agreement and 
the chance correlated measure, Gwet's AC1/gamma coefficient \citep{gwetac1} as the  inter-rater agreement. Gwet's AC1/gamma coefficient overcomes the issue where traditional measures, such as Cohen's Kappa, are not robust to skewed distributions of  rankings.
Figure~\ref{fig:interface} shows the interface for human evaluation study.

\section{Hyper-parameter search domain for directed generation}
\label{append:unlikeforsumm}
During decoding, we apply length normalization following previous works. For the hyper-parameter in length normalization (beam search decoding), we use $\beta \in \{0.0, 0.5, 1.0, 1.5, 2.0\}$ for text summarization and $\beta \in \{0.0, 0.1, 0.2, 0.3, 0.4, 0.5, 0.6, 0.7, 0.8,0.9,1.0\}$ for image paragraph captioning.

\section{Experimental results on open-ended generation}

\subsection{Full experimental results on WikiText-103}
\label{append:fullresultsopen}

We present the full experimental results on WikiText-103 \citep{merity2016pointer} test set for open-ended generations in Table~\ref{table:openendedfull}. All the numbers are averaged over 3 runs with different random seeds and shown together with standard deviations.

\begin{table*}[th]
\caption{Results for open-ended generations on the {\textbf{Wikitext-103 testset}}.
\textbf{ppl}, \textbf{uniq} and \textbf{Rep/l} are computed at BPE-level and the rest are at word-level. The  ``$\uparrow$'' denotes higher value for better performance and ``$\downarrow$'' is the opposite. {Number marked with * are estimated based on the testset.}}
\centering
\scalebox{0.76}{\begin{tabular}  {lccccccccc} 
\toprule
 &  \multicolumn{5}{c}{\textbf{Language Modeling}}
 & \multicolumn{4}{c}{\textbf{Auto Completion}} \\
 \cmidrule(lr){2-6}  \cmidrule(lr){7-10} 

\textbf{Models} & \bf{ppl} $\downarrow$ & \bf{uniq} $\uparrow$ &\bf{Rep/16} $\downarrow$  &\bf{Rep/32} $\downarrow$  & \bf{Rep/128} $\downarrow$ &\bf{Rep-1} $\downarrow$ & \bf{Rep-2} $\downarrow$ & \bf{Rep-3} $\downarrow$  & \bf{uniq-w} $\uparrow$\\
\midrule

MLE             & \bf{13.24}$_{\pm2e-4}$ & 12.54k$_{\pm4e-3}$ & 0.234$_{\pm5e-6}$ & 0.380$_{\pm8e-6}$ & 0.619$_{\pm7e-6}$ & 0.661$_{\pm1e-5}$ & 0.500$_{\pm3e-5}$ & 0.424$_{\pm7e-5}$  & 16.83k$_{\pm1e-1}$\\

UL ($\alpha=1.0$)    & 16.06$_{\pm2e-2}$ & 13.18k$_{\pm6e-3}$ & 0.212$_{\pm1e-6}$ & 0.341$_{\pm1e-7}$ & 0.558$_{\pm9e-6}$ & 0.559$_{\pm6e-5}$ & 0.363$_{\pm2e-4}$ & 0.291$_{\pm3e-4}$ & 19.11k$_{\pm7e-2}$  \\

SG ($\gamma=0.2$)  & 14.20$_{\pm2e-2}$ & \bf{13.61k}$_{\pm2e-3}$ & \bf{0.197}$_{\pm6e-7}$ & \bf{0.317}$_{\pm1e-6}$ & \bf{0.522}$_{\pm4e-6}$ & \bf{0.443}$_{\pm9e-7}$ & \bf{0.215}$_{\pm2e-6}$ & \bf{0.143}$_{\pm4e-6}$  & \bf{22.25k}$_{\pm2e-2}$ \\

\midrule
Human                & - & 18.27k & 0.177 & 0.285 & 0.480 & {0.382*} & {0.096*} & {0.037*} & {27.55k*}\\
\bottomrule
\end{tabular}}
\label{table:openendedfull}
\end{table*}

In addition, we provide the full results \wrt different hyper-parameters for UL and SG on the WikiText-103 validation set in Table~\ref{table:openendedonvalid}.  
\begin{table*}[t!]
\caption{Results for open-ended generation tasks on the \textbf{Wikitext-103 validation set}.
\textbf{ppl}, \textbf{uniq} and \textbf{Rep/l} are computed at BPE-level and the rest are at word-level. The  ``$\uparrow$'' denotes higher value for better performance and ``$\downarrow$'' is the opposite. {Number marked with * are estimated based on the testset.} {The results are averaged over 3 runs with different random seeds.}}
\centering
\scalebox{0.97}{\begin{tabular}  {lccccccccc} 
\toprule
 &  \multicolumn{5}{c}{\textbf{Language Modeling}}
 & \multicolumn{4}{c}{\textbf{Auto Completion}} \\
 \cmidrule(lr){2-6}  \cmidrule(lr){7-10} 

\textbf{Models} & \bf{ppl} $\downarrow$ & \bf{uniq} $\uparrow$ &\bf{Rep/16} $\downarrow$  &\bf{Rep/32} $\downarrow$  & \bf{Rep/128} $\downarrow$ &\bf{Rep-1} $\downarrow$ & \bf{Rep-2} $\downarrow$ & \bf{Rep-3} $\downarrow$  & \bf{uniq-w} $\uparrow$\\
\midrule
MLE	 & 13.17 &	12.52k&	0.236&	0.384&	0.621&	0.665&	0.510&	0.428&	16.71k \\
UL($\alpha=0.5$)&	14.91&	12.45k&	0.217&	0.350	&0.579&	0.601&	0.424 &	0.348&	18.02k\\
UL($\alpha=1.0$)	&16.52&	12.77k&	0.210&	0.336&	0.552&	0.551&	0.359&	0.289&	19.14k\\
UL($\alpha=1.5$)	&19.63&	13.41k&	0.201&	0.315&	0.523&	0.489&	0.267&	0.205&	22.00k\\
SG($\gamma=0.2$) &	14.43&	13.73k&	0.195&	0.316&	0.518&	0.451&	0.237&	0.175&	22.29k \\
SG($\gamma=0.5$) & 13.53&	13.25k&	0.218&	0.352&	0.576&	0.561&	0.389&	0.331&	19.13k \\
SG($\gamma=0.8$) &13.27&	12.79k&	0.229&	0.369&	0.603&	0.625&	0.443&	0.365&	17.59k\\
\midrule
Human&	--	&17.68k	&0.173	&0.278&	0.470&	0.376&	0.097&	0.032	&27.63k\\
\bottomrule
\end{tabular}}
\label{table:openendedonvalid}
\end{table*}

\subsection{Open-ended generations results on IMDB dataset}
\label{append:imdb}
Table~\ref{table:openendedonimdb} shows the open-ended generation results on movie revies from IMDB dataset. 

\begin{table*}[ht]
\caption{Results for open-ended generations on movie reviews from \textbf{IMDB} dataset.
\textbf{ppl}, \textbf{uniq} and \textbf{Rep/l} are computed at BPE-level and the rest are at word-level. {Numbers marked with * are estimated based on the movie reviews from IMDB.}}
\centering
\scalebox{0.83}{\begin{tabular}  {lccccccccc} 
\toprule
 &  \multicolumn{5}{c}{\textbf{Language Modeling}}
 & \multicolumn{4}{c}{\textbf{Auto Completion}} \\
 \cmidrule(lr){2-6}  \cmidrule(lr){7-10} 

\textbf{Models} & \bf{ppl} & \bf{uniq}  &\bf{Rep/16}  &\bf{Rep/32}   & \bf{Rep/128}  &\bf{Rep-1}  & \bf{Rep-2}  & \bf{Rep-3}   & \bf{uniq-w} \\
\midrule

MLE & {100.764} &  7.48k  & 0.153 & 0.254 &   0.449     & 0.662 & 0.499 & 0.429 & 7.70k\\

UL ($\alpha=1.0$) & 108.334 & 8.09k & 0.123 & {0.205} & {0.373} & 0.545 & 0.346 & 0.274 & 9.31k     \\

SG ($\gamma=0.2$) & 110.451 & {8.14k} & {0.114} & 0.187 & 0.344 & {0.383} & {0.142} & {0.081} & {10.42k} \\

\midrule
Human  & - & 14.49k &  0.118 & 0.208 & 0.378 & 0.329* & 0.084* & 0.009* & *19.11k\\
\bottomrule
\end{tabular}}
\label{table:openendedonimdb}
\end{table*}

\section{Experimental details}
In this section, we present the details of the datasets used in our experiments as well as the necessary experimental setup.
All the experiments were conducted with a single GPU on our machine (CPU: Intel(R) Xeon(R) Gold 6240 CPU @ 2.60GHz; GPU: NVIDIA RTX 2080Ti). 

For each task in our experiments, we use the same model architecture and train it with different objectives (\ie MLE, {ScaleGrad} and unlikelihood). 
The hyper-parameters that are used for different training objectives in the same task are exactly same, except for the ones described in Appendix~\ref{append:unlikeforsumm}.
We list the key hyper-parameters in this section.

\label{append:hyperparameters}
\subsection{Open-ended generation}

\paragraph{Dataset} The WikiText-103 \citep{merity2016pointer} is a collection of over 100 million tokens extracted from the set of verified Good and Featured articles on Wikipedia. The training, validation and test sets contain 104m, 218k and 245k tokens, respectively.

\paragraph{Experiments} For all the experiments, we use the same setup and the same hyper-parameters as listed in Table~\ref{table:hyper_open}, except for the method-specific hyper-parameters. We load the GPT-2 medium and fine-tune it on WikiText-103 with a maximum of 35k iterations and select the model based on the validation perplexity.

\begin{table}[ht]
    \centering
     \caption{Hyper-parameters for open-ended generation. \textbf{M} denotes the model-specific hyper-parameters. \textbf{lr$_0$} is initial learning rate.}
    \label{table:hyper_open}
    \begin{tabular}{l c c c  }
    \\
    \toprule
    \bf{Models} & \bf{lr$_0$} & \bf{M} & \bf{batch} \\ 
    \midrule
    MLE & \num{2e-5} & -- & 300 \\
    UL    & \num{2e-5} & 0.5/1.0/1.5 & 300\\
    ScaleGrad & \num{2e-5} & 0.2/0.5/0.8 & 300\\
    \bottomrule
    \end{tabular}
\end{table}

\subsection{Summarization}
\paragraph{Dataset} We use CNN/DM \citep{hermann2015CNN,nallapati-etal-2016-abstractive} and NYT50 \citep{durrett-etal-2016-learning} in our experiments for text summarization. Table~\ref{table:dataset_statssumm} shows the dataset statistics in details.
\begin{table}[ht]
    \centering
    \caption{Dataset statistics for summarization.}
    \label{table:dataset_statssumm}
     \scalebox{0.86}{\begin{tabular}{ l ccc } 
         \\
         \toprule
         \bf{Dataset}  & \bf{Training Size} & \bf{Validation Size} & \bf{Test Size}\\ 
         \midrule
         \bf{CNN/DM}  & 287,227 & 13,368 & 11,490 \\
          \bf{NYT50}   & 96,834 & 4,000 & 3,452 \\
        \bottomrule
    \end{tabular} }                  
\end{table}

\paragraph{Experiments} 
The models are taken from \citep{liu-lapata-2019-text} and we train the models for the abstractive summarization with MLE, unlikelihood training and {ScaleGrad} on CNN/DM and NYT50. We list the hyper-parameters that we used in Table~\ref{table:hyper_summ}.

\begin{table}[ht]
    \centering
     \caption{Hyper-parameter lists for text summarization. \textbf{M} denotes the model-specific hyper-parameters. \textbf{lr$^\text{BERT}_0$} and \textbf{lr$^\text{dec}_0$} stand for initial learning rate for BERT and Transformer decoder. $\beta$ is the hyper-parameter in length normalization.
     }
    \label{table:hyper_summ}
    \scalebox{0.86}{\begin{tabular}{l c c c c c c}
    \\
    \toprule
    \bf{Models} & \bf{lr$^\text{BERT}_0$} & \bf{lr$^\text{dec}_0$} & \bf{M} & \bf{batch} &  \bf{$\beta$} & \bf{Beam Size}\\ 
    \midrule
    \bf{CNN/DM}\\
    MLE & 0.002 & 0.2 & -- & 140 & 1.0 & 5\\
    UL    & 0.002 & 0.2 & 0.5 & 140 & 2.0 & 5\\
    ScaleGrad & 0.002 & 0.2 & 0.8 & 140 & 1.5 & 5\\
    \midrule
    \bf{NYT50}\\
    MLE & 0.002 & 0.2 & -- & 140 & 1.5 & 5\\
    UL    & 0.002 & 0.2 & 0.5 & 140 & 2.0 & 5\\
    ScaleGrad & 0.002 & 0.2 & 0.8 & 140 & 1.5 & 5\\
    \bottomrule
    \end{tabular}}
\end{table}

\subsection{Image paragraph generation}
\paragraph{Dataset} We use the image paragraph captioning corpus Visual Genome dataset, introduced by \citet{krause2016paragraphs}. The dataset contains 14,575 training, 2,487 validation, and 2,489 testing images. The average length of description paragraph is 67.50 tokens.

\paragraph{Experiments} We follow the same experimental setup as in \citep{melas-kyriazi-etal-2018-training}. We train the model with different objectives and choose the model for testing based on the validation loss. During generation, tri-gram blocking and length-normalization are applied. Hyper-parameters that are used in our experiments are listed in Table~\ref{table:hyper_imgp}.

\begin{table}[ht]
    \centering
     \caption{Hyper-parameter lists for image paragraph captioning. \textbf{M} denotes the model-specific hyper-parameters. \textbf{lr$_0$} is initial learning rate.}
    \label{table:hyper_imgp}
    \scalebox{0.83}{\begin{tabular}{l c c c c }
    \\
    \toprule
    \bf{Models} & \bf{lr$_0$} & \bf{M} & \bf{batch} & $\beta$ (w/o \& w/ 3-blocking)\\ 
    \midrule
    MLE & \num{5e-4} & -- & 10 & 0.0/0.2\\
    UL    & \num{5e-4} & 0.5 & 10 & 0.0/0.3\\
    ScaleGrad & \num{5e-4} & 0.5 & 10 & 0.6/0.6\\
    \bottomrule
    \end{tabular}}
\end{table}

\section{Experimental results of different decoding strategies for auto-completion. }
\label{append:decodingstrategy}

\begin{table}[ht]
     \caption{Results of different decoding strategies for auto-completion.}
    \centering
    \label{table:appendix decodingmethods}
    \scalebox{0.76}{\begin{tabular} {lccccc}
    \\ 
    \toprule
    \bf{Approaches} & \bf{ppl} &\bf{Rep-1} & \bf{Rep-2} & \bf{Rep-3} & \bf{uniq-w} \\ 
    \midrule
    \textbf{ScaleGrad} \\
    Greedy Search ($\gamma=0.2)$& 14.20 & {0.443} & {0.215} & {0.144}  & {22.25k}\\
    Beam Search (b = 3) & 14.20 &  0.422 & 0.210 & 0.134 & 8.75k\\
    Beam Search (b = 6) & 14.20&  0.453 & 0.250 & 0.171 & 8.32k\\
    Beam Search (b = 10) & 14.20&  0.489 & 0.298 &  0.214 & 8.00k\\
    Top-$p$ (p = 0.3) & 14.20&  0.356 & 0.107 & 0.049 & 30.48k\\
    Top-$p$ (p = 0.9) & 14.20&  0.217 & 0.027 & 0.008 & 52.76k \\
    Top-$k$ (k = 40) & 14.20&    0.254 & 0.039 & 0.012 & 39.50k \\
    Top-$k$ (k = 100) & 14.20&  0.234  &  0.031 & 0.010 & 44.27k \\
    \midrule
    \textbf{UL}\\
    Greedy Search ($\alpha=1.0$)&	16.06&	0.559&	0.363&	0.291&	19.11k\\
    Beam Search ($b=6$)&	16.06&	0.577&	0.418&	0.325&	7.49k\\
Top-$p$ ($p=0.3$)&	16.06&	0.444&	0.176&	0.070&	24.45k\\
Top-$k$ ($k=40$)	& 16.06	&0.336&	0.067&	0.021&	31.89k\\
\midrule
    \textbf{MLE} \\
    MLE	& 13.24&	0.661&	0.500&	0.424&	16.83k\\
Beam Search ($b=6$) &	13.24&	0.697&	0.566&	0.463&	6.11k\\
Top-$p$ ($p=0.3$)&	13.24&	0.558&	0.210&	0.116&	20.13k\\
Top-$k$ ($k=40$) &	13.24&	0.485&	0.154&	0.076&	24.26k\\
    \midrule
    Human	&--	&0.382&	0.096&	0.037&	27.55k\\
    \bottomrule
    \end{tabular}}
\end{table}
Table~\ref{table:appendix decodingmethods} shows the results for the  auto-completion task when we train the model with {ScaleGrad} and infer with different decoding strategies.

\section{{Stochastic decoding for image paragraph captioning}}
\label{append:stochastic for img}
We apply different stochastic decoding strategies for the MLE baseline on image paragraph captioning and report the results in Table~\ref{table:toppkonimg}. The experimental results demonstrate that stochastic decoding strategies do not work well in directed generation tasks, which is consitent with our findings in summarizaiton experiments.

\begin{table}[ht]
    \centering
    \captionof{table}{Image paragraph captioning results for stochastic decoding on Visual Genome testset.}
   \label{table:toppkonimg}
    \scalebox{1}{\begin{tabular}{lc}
    \toprule
    \bf{Models} & \bf{CIDEr} \\
    \midrule
    Top-$p$ (p=0.3) & 19.54  \\
    Top-$p$ (p=0.9) & 18.67  \\
    Top-$k$ (k=40) & 18.73 \\
    Top-$k$ (k=100) \ \quad \quad \ & 18.05 \\
    \midrule
    MLE w/ 3-block & 22.77 \\
    \bottomrule
    \end{tabular}}
\end{table}

\section{Hyper-parameter sensitivity}
\label{append:hyper-sensi}
\begin{figure*}[ht]
     \centering
     \begin{subfigure}[b]{0.32\textwidth}
         \centering
         \includegraphics[width=\textwidth]{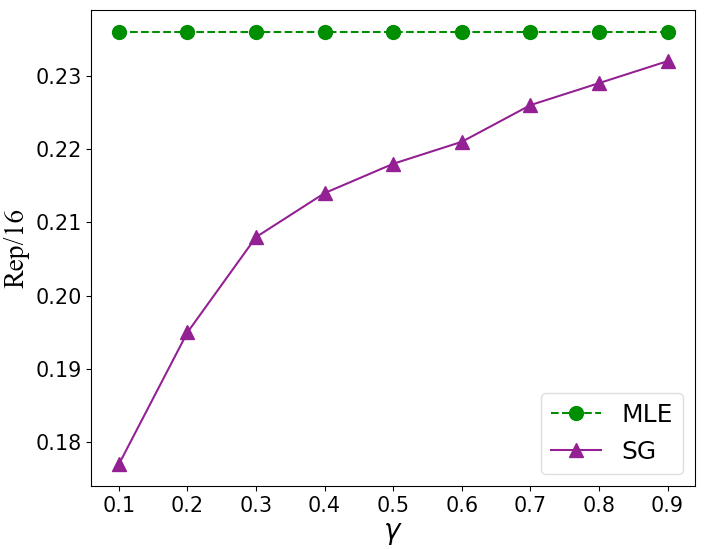}
         \caption{Rep/$16$}
         \label{fig:rep16}
     \end{subfigure}
     \hfill
     \begin{subfigure}[b]{0.32\textwidth}
         \centering
         \includegraphics[width=\textwidth]{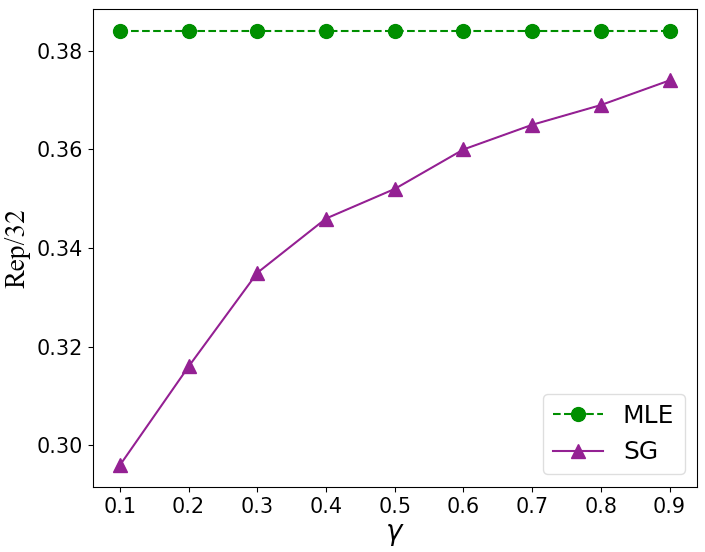}
         \caption{Rep/$32$}
         \label{fig:rep32}
     \end{subfigure}
     \hfill
     \begin{subfigure}[b]{0.32\textwidth}
         \centering
         \includegraphics[width=\textwidth]{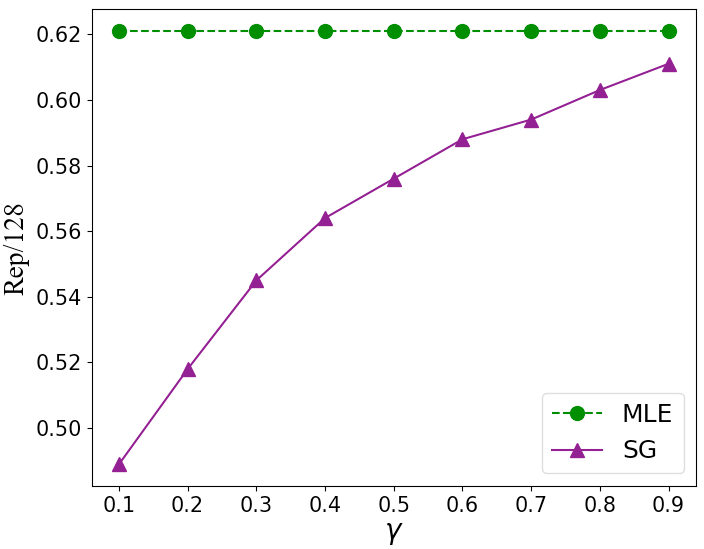}
         \caption{Rep/$128$}
         \label{fig:rep128}
     \end{subfigure}
        \vspace{-0.5em}
        \caption{Hyper-parameter ($\gamma$) sensitivity in the language modeling task {on Wikitext-103 development set}. }
        \label{fig:detail graphs}
        \vspace{-1em}
\end{figure*}

To fully present the sensitivity of Rep/$l$ to the hyper-parameter, we further show how the Rep/$l$ (\ie $l$=16, 32 and 128) change with $\gamma$ in Figure~\ref{fig:detail graphs}.

\section{Examples}
\label{append:examples}
In the following, we show the examples of generated texts in three tasks: auto-completion (Table~\ref{table:example auto-completion} and Table~\ref{table:example2 auto-completion}), image paragraph captioning (Table~\ref{table:example1 img-para} and Table~\ref{table:example2 img-para}) and text summarization (Table~\ref{table:example1 summarization}, Table~\ref{table:example2 summarization}, Table~\ref{table:example3 summarization} and Table~\ref{table:example4 summarization}). In addition, Table~\ref{table:example ptb} and Table~\ref{table:example imdb} show the example of auto completion on PTB testset and movie reviews from IMDB dataset.

\begin{table*}[ht]
    \centering
    \caption{Example 1 of generated continuations for different training objectives.}
    \label{table:example auto-completion}
    \begin{tabular}{p{0.97\textwidth}}
        \toprule
        \textbf{Prefix} \\
        \midrule
        Robert Boulter is an English film, television and theatre actor. He had a guest  starring role on the television series The Bill in 2000. This was followed by a starring\\
        \midrule
         \textbf{MLE}  \\
         \midrule
         role in the film The Last of the Mohicans in 2002. He has appeared in the television series The X Factor, The X Factor Australia, The X Factor New Zealand, The X Factor South Africa, The X Factor New Zealand, The X Factor South Africa, The X Factor New Zealand, The X Factor New Zealand, The X Factor New Zealand, The X Factor New Zealand, The X Factor New Zealand, The X Factor New Zealand, The X Factor New Zealand, The X Factor New Zealand \\
         \midrule
         \textbf{Unlikelihood}  \\
         \midrule
         role in the film The Last of the Mohicans, which was released in 2002. He also appeared in the television series The X Factor, where he played the role of a DJ. In 2005, he starred in the film The Last of the Mohicans, which was released in 2006. He also appeared in the television series The X Factor, where he played the role of a DJ. In 2008, he starred in the television series The X Factor, where he played the role \\
         \midrule
         \textbf{ScaleGrad}  \\
         \midrule
         role in the BBC's comedy series The Office in 2002. In 2003, he appeared in the BBC's crime drama series The Yard, playing a police officer who is caught up in a murder investigation. He also appeared in the BBC's crime drama series The X Factor, playing a man who is accused of murdering his wife.  In 2004, he starred in the BBC's comedy series The Office, playing a fictional version of himself. He has appeared in several \\
         \bottomrule
    \end{tabular}
    
\end{table*}

\begin{table*}[ht]
    \centering
    \caption{Example 2 of generated continuations for different training objectives.}
    \label{table:example2 auto-completion}
    \begin{tabular}{p{0.97\textwidth}}
        \toprule
        \textbf{Prefix} \\
        \midrule
        You's patriotism, and Mei's reflections on the quotidian are a few examples. More broadly, Du Fu's work in transforming the lushi from mere word play into ``a vehicle for serious poetic utterance" \\
        \midrule
         \textbf{MLE}  \\
         \midrule
         is a major contribution to the genre.  The first of the three major works of the genre,  was written in the early 1920s, and was the first of the three to be published in English. It was published in the United States in the United States, and was translated into French in the 1930s. The work was the first of the three\\
         \midrule
         \textbf{Unlikelihood}  \\
         \midrule
         is a major influence on the modern Chinese novel. The modern Chinese novel is often characterized as being ``literary in nature", but it is not limited to literary genres. The modern Chinese novel has been described as ``a literary genre that has been influenced by the Western literary tradition". The modern Chinese novel has been described as ``a literary genre that has been influenced by the Western literary tradition"\\
         \midrule
         \textbf{ScaleGrad}  \\
         \midrule
         is a powerful example of how the Chinese literary tradition can be used to explore themes of nationalism and self determination. In addition to his own poetry, Du Fu wrote several works of prose. The first was a collection of short stories called ``The Story of the Three Sisters", which was published in 1885. It was followed by a series of novels, including the first two volumes of the Song dynasty epic poem ``The Three Sisters"\\
         \bottomrule
    \end{tabular}
    
\end{table*}

\begin{table*}[ht]
    \centering
    \caption{Example 1 of generated paragraphs for different training objective on Visual Genome testset.}
    \label{table:example1 img-para}
    \begin{tabular}{p{0.97\textwidth}}
        \toprule
        \textbf{Reference}\\
        \midrule
        Three adult horses are in a field and a small horse. A horse color light brown has the neck low and is eating grass. This horse has the mane and the tail color tan. Two dark brown horses also are eating grass from the ground. The baby horse has a white spot on the front and is tiny. On back there is a wall of stones.\\
        \midrule
         \textbf{MLE w/o tri-gram blocking}  \\
         \midrule
         There are two horses in a field. The horses are brown and white. The horses are brown and white. The horses are brown and white. The horses are brown and white. The horses are brown and white. The horses are brown and white. The horses are brown. The horses are brown. The horses are brown. The horses are brown. The horses are brown. The horses are brown. The horses are brown. The horses are brown. The horses are brown . The horses are brown. The horses are brown. The horses are brown. The horses are brown. The horses are brown. The horses are brown. \\
         \midrule
         \textbf{MLE w/ tri-gram blocking}  \\
         \midrule
         There are two horses in a field. The horses are brown and white. The horse on the right is brown and brown. The horse is brown. The other horse is white. The grass is green and green. The sky is very and green. There are trees in the background. The trees are green and healthy. The horses are black. The sun is shining brightly. \\
         \midrule
         \textbf{ScaleGrad w/ tri-gram blocking}  \\
         \midrule
         Two horses are standing in a field. The field is covered with green grass and weeds. The horse is brown and white. The horses are brown and white. There is a large tree behind the horses. The sky is blue and clear. There are many trees in the background. The grass is green. The sun is shining brightly. The sky is mostly cloudy. \\
         \bottomrule
    \end{tabular}
    
\end{table*}

\begin{table*}[ht]
    \centering
    \caption{Example 2 of generated paragraphs for different training objectives on Visual Genome testset.}
    \label{table:example2 img-para}
    \begin{tabular}{p{0.97\textwidth}}
        \\
        \toprule
        \textbf{Reference}\\
        \midrule
        There is a white and yellow train with red doors next to a red and yellow train. They are on top of the railroad tracks. There is a brick walkway next to the trains.\\
        \midrule
         \textbf{MLE w/o tri-gram blocking}  \\
         \midrule
         A train is on the tracks. The train is yellow and yellow. The train is red and white . The train is red and white. The train is red and white . The train is on the train. The train is yellow and yellow. The train is blue and white . The train is on the tracks. The train is yellow and yellow . The train is blue and white. The train is blue and white. The train is on the tracks . The train is yellow. The train is blue. The train is blue. The train is blue. The train is blue. The train is blue. The train is blue. The train is blue. The train is blue. The train is blue. The train is blue.\\
         \midrule
         \textbf{MLE w/ tri-gram blocking}  \\
         \midrule
         A train is on the tracks. The train is yellow and yellow. The train has a yellow stripe on the front. The front of the train is red. The sky is blue. There are many windows on the train. There is a train on the platform. The platform is made of metal. The is a platform on the side of the train. The are many tracks on the train. There are wires on the ground. There is a building behind the train tracks. There is a large building behind the train.\\
         \midrule
         \textbf{ScaleGrad w/ tri-gram blocking}  \\
         \midrule
         A train is on the tracks. There are two sets of tracks next to the train. The train is white and yellow. There is a large white building behind the trains.\\
         \bottomrule
    \end{tabular}
    
\end{table*}

\begin{table*}[ht]
    \centering
    \caption{Example 1 of summaries generated by the model trained with different training objective on CNN/DM testset.}
    \label{table:example1 summarization}
    \begin{tabular}{p{0.97\textwidth}}
        \toprule
        \textbf{Reference} \\
        \midrule
        report by dick marty, lead investigator for council of europe, gives bleak description of secret prisons run by cia in poland and romania from 2003 to 2006. marty says he relied heavily on information from cia agents. critics in poland and romania attack marty's use of anonymous sources.\\
        \midrule
         \textbf{MLE}  \\
         \midrule
         swiss sen dick marty reports that central intelligence agency operates secret prisons run by american in poland and romania from 2003 to 2006. says prison were operated exclusively by americans in poland and.\\
         \midrule
         \textbf{Unlikelihood}  \\
         \midrule
         swiss sen dick marty reports that secret prisons run by central intelligence agency in eastern europe, with information he says is gleaned from anonymous intelligence agents. report is prepared by swiss senator investigating cia operations for council of europe, 46 - nation rights group. scathing report says prison were operated exclusively by americans in poland and romania from 2003 to 2006.\\
         \midrule
         \textbf{{ScaleGrad}}  \\
         \midrule
        dick marty, swiss senator investigating cia operations for council of europe, gives bleak description of secret prisons run by central intelligence agency in eastern europe, with information he says is gleaned from anonymous intelligence agents. report says prisons were operated exclusively by americans in poland and romania from 2003 to 2006.\\
         \bottomrule
    \end{tabular}
    
\end{table*}

\begin{table*}[ht]
    \centering
    \caption{Example 2 of summaries generated by the model trained with different training objective on CNN/DM testset.}
    \label{table:example2 summarization}
    \begin{tabular}{p{0.97\textwidth}}
        \toprule
        \textbf{Reference} \\
        \midrule
        middleton wrote about eating whale for a national newspaper travel story. she described eating it at the \#123-a-night juvet landscape hotel. whale and dolphin conservation group said the news was `disappointing'. the wdc said ms.middleton should have gone whale watching instead.\\
        \midrule
         \textbf{MLE}  \\
         \midrule
         the 31-year-old has been slammed for `promoting the cruel and unnecessary whaling industry'. the 31-year-old boasted of eating minke whale in norway, to the horror of environmental campaigners who say she is helping promote an `unimaginably cruel' trade. the hunting and importation of whale meat is banned in britain , as it is in most parts of the world.\\
         \midrule
         \textbf{Unlikelihood}  \\
         \midrule
        the 31-year-old boasted of eating minke whale at juvet landscape hotel in norway. the 31-year-old has been slammed for `promoting the cruel and unnecessary whaling industry '<q>environmental campaigners say she is helping promote an ` unimaginably cruel ' trade.\\
         \midrule
         \textbf{{ScaleGrad}}  \\
         \midrule
        duchess of cambridge's brother-in-law has led a personal crusade against poaching and protecting wildlife. pippa middleton boasted of eating minke whale in norway, conservation group said she is helping promote `promoting the cruel and unnecessary whaling industry'. \\
         \bottomrule
    \end{tabular}
    
\end{table*}

\begin{table*}[ht]
    \centering
    \caption{Example 1 of summaries generated by the model trained with different training objective on NYT50 testset.}
    \label{table:example3 summarization}
    \begin{tabular}{p{0.97\textwidth}}
        \toprule
        \textbf{Reference} \\
        \midrule
        protesters angry over bribery scandal involving state-run oil company petrobras. brazilian president dilma rousseff also is struggling with an economic downturn.\\
        \midrule
         \textbf{MLE}  \\
         \midrule
         protesters are calling for president dilma rousseff to be impeached. rousseff's supporters call for the president to be impeachment<q>they say there hasn't been any evidence she was involved in the corruption scandal.\\
         \midrule
         \textbf{Unlikelihood}  \\
         \midrule
         protesters are calling for president dilma rousseff to be impeached. there is a number of issues at play. one of the biggest is an investigation into a multimillion-dollar kickback scheme at the state-run oil company petrobras.\\
         \midrule
         \textbf{ScaleGrad}  \\
         \midrule
        president dilma rousseff is facing an economic downturn and a massive bribery scandal. rousseff has defended her right to protest and acknowledged the need to clean up corruption at petrobras.\\
         \bottomrule
    \end{tabular}
    
\end{table*}

\begin{table*}[ht]
    \centering
    \caption{Example 2 of summaries generated by the model trained with different training objective on NYT50 testset.}
    \label{table:example4 summarization}
    \begin{tabular}{p{0.97\textwidth}}
        \toprule
        \textbf{Reference} \\
        \midrule
        wollemi pine, which flourished 200 million years ago, is available to gardeners and can be grown indoors or outdoors. thought to extinct, it was discovered in 1994 in wollemi national park in australia. botanists took cuttings from wild colony of trees and began propagating baby pines and sending them to botanic gardens around world.\\
        \midrule
         \textbf{MLE}  \\
         \midrule
         new york city plant - o - ramachan pine, which grew among dinosaur 200 million years ago, is named after dr david noble, who rappelled into deep gorge in blue mountains of wollemi national park, ten miles west of sydney, australia. photos. map. pine was discovered until 1994, when bush delay and national parks officer david noble tried rappelled into woods of wollemi, which he had never seen before. scientists had evidence of wollemi pine's existence, but only from fossils.\\
         \midrule
         \textbf{Unlikelihood}  \\
         \midrule
         nicholas d kristof op-ed column on plight of wollemi pine, one of oldest on earth who grew among dinosaur 200 million years ago. notes that it was not discovered until 1994, when bush rescue officer david noble rappelled into deep gorge in blue mountains of sydney, australia, and came back up with branch of tall tree that he never saw before.\\
         \midrule
         \textbf{ScaleGrad} \\
         \midrule
        nicholas d kristof op-ed column on new species of wollemi pine, which grew among dinosaur 200 million years ago and was discovered until 1994, when bush delay and national parks officer david noble rappelled into deep gorge in blue mountains of wollemi national park, 120 miles west of sydney, australia, and discovered new species, wollemia nobilis, named after noble. \\
         \bottomrule
    \end{tabular}
    
\end{table*}



\begin{table*}[ht]
    \centering
    \caption{Example of generated continuations for different training objectives on PTB testset.}
    \label{table:example ptb}
    \begin{tabular}{p{0.97\textwidth}}
        \toprule
        \textbf{Prefix} \\
        \midrule
         last resort who were criticized after the 1987 crash -- once again couldn't handle the selling pressure. Big investment banks refused to step up to the plate to support the beleaguered floor traders by buying big blocks of stock, traders say. Heavy selling\\
        \midrule
         \textbf{MLE}  \\
         \midrule
         pressure was also felt by the Wall Street banks. The Wall Street banks were forced to sell more stock to hedge funds and other investors, and to buy more stock to hedge against losses. The Wall Street banks were forced to sell more stock to hedge funds and other investors, and to buy more stock to hedge against losses. The Wall Street banks were forced to sell more stock to hedge funds and other investors, and to buy more stock to hedge against losses.\\
         \midrule
         \textbf{Unlikelihood}  \\
         \midrule
          pressure was also felt by the banks themselves. In 1987, the New York Federal Reserve Bank, which had been the largest bank in the United States since the 1930s, was forced to sell \$1.5 billion worth of stock. The bank's stock price fell to \$0.001, and it was forced to close its New York branch. The bank's stock price fell to \$0.001, and it was forced to close its New\\
         \midrule
         \textbf{ScaleGrad}  \\
         \midrule
         by the big banks led to a sharp drop in prices, and the market collapsed. In 1992, the Federal Reserve Bank of New York issued a report on the financial crisis, which concluded that the Fed's actions had not been sufficient to prevent the collapse of the housing market. The report recommended that the Fed increase its balance sheet by \$1.5 trillion (equivalent to about \$2.5 trillion today), but this was not implemented.\\
         \bottomrule
    \end{tabular}
    
\end{table*}

\begin{table*}[ht]
    \centering
    \caption{Example of generated continuations for different training objectives on movie reviews from IMDB dataset.}
    \label{table:example imdb}
    \begin{tabular}{p{0.97\textwidth}}
        \toprule
        \textbf{Prefix} \\
        \midrule
         One of the other reviewers has mentioned that after watching just 1 Oz episode you'll be hooked. They are right, as this is exactly what happened with me. The first thing that struck me about Oz was its brutality and unflinching scenes of violence. \\
        \midrule
         \textbf{MLE}  \\
         \midrule
         I was not expecting anything like that in a show about a family. I was expecting a show about a family, but I was expecting something more. I was expecting a show about a family, but I was expecting something more. I was expecting a show about a family, but I was expecting something more. I was expecting a show about a family, but I was expecting something more. I was expecting a show about a family, but I was expecting something more. I was expecting a show\\
         \midrule
         \textbf{Unlikelihood}  \\
         \midrule
         I was immediately hooked. The violence was so graphic that I was completely taken aback by how it was handled. I was also struck by the fact that the show was so well-written that it was almost impossible to not be entertained. The second thing that struck me about Oz was its ability to make me laugh. I was so taken aback by the violence that I couldn 't help but laugh. The violence was so graphic that I was completely taken aback by how it \\
         \midrule
         \textbf{ScaleGrad}  \\
         \midrule
         It was also very funny, but not funny enough to make me laugh out loud. The second thing that struck me was the fact that the show's characters were so well developed that they were able to carry their own weight without being annoying or annoying.   In its original broadcast, ``Oz" was watched by 9.5 million viewers, according to Nielsen ratings. It was the highest rated\\
         \bottomrule
    \end{tabular}
    
\end{table*}



\end{document}